\documentclass[10pt]{article} 

\usepackage[accepted]{rlj} 
\usepackage{amsthm}

\usepackage{amsthm}
\usepackage{tcolorbox}

\newtheorem{theorem}{Theorem}

\newtheorem{proposition}{Proposition}
\usepackage{amssymb}            
\usepackage{mathtools}          
\usepackage{mathrsfs}           
\usepackage{graphicx}
\usepackage{pgffor}
\usepackage{pgfplots}
\usepackage{graphicx}           
\usepackage{subcaption}         
\usepackage[space]{grffile}     
\usepackage{url}                
\usepackage{lipsum}             
\usepackage{subcaption} 

\title{Adaptive Multi-Horizon Reinforcement Learning}

\setrunningtitle{Adaptive Multi-Horizon Reinforcement Learning}

\author{Manoosh Samiei\textsuperscript{1,2}, Doina Precup\textsuperscript{1,2}, Paul Masset\textsuperscript{2,3}}

\emails{manoosh.samiei@mail.mcgill.ca,
doina.precup@mcgill.ca, paul.masset@mcgill.ca}
\affiliations{
$^{1}$\textbf{Department of Computer Science, McGill University, Canada}\\
$^{2}$\textbf{Mila - Quebec Artificial Intelligence Institute}\\
$^{3}$\textbf{Department of Psychology, McGill University, Canada}}

\par 

\contribution{
We study the role of the discount factor across environments with varying reward sparsity, spatial structure, and temporal horizons, showing that optimal discounting is dependent on the environment and task. We also demonstrate that different discount factors induce different greedy policies.
}
{
None
}

\contribution{
Inspired by dopaminergic circuits and their interactions with the prefrontal cortex, we propose a mixture-of-experts Q-learning framework that adaptively combines multiple $\gamma$-discounted action-value functions through a state-dependent mixture network, enabling the dynamic selection of effective temporal horizons.
}
{
None
}
\keywords{Discounting, Multi-Timescale Reinforcement Learning, Reward Prediction Error, Dopamine, Horizon, Continual Learning} 

\summary{
We introduce a mixture-of-experts reinforcement learning framework that dynamically adapts temporal discounting by combining multiple action-value functions trained with different discount factors. Each expert ($Q_i(s,a)$) corresponds to a fixed discount factor ($\gamma_i$), capturing value estimates over different temporal horizons. These experts are learned independently using tabular Expected SARSA.

A state-dependent gating network learns to combine these experts into a single mixed action-value function ($Q_{\text{mix}}(s,a)$) via a softmax-weighted linear mixture. The gating mechanism is trained using average-reward differential TD learning, enabling it to adaptively select effective temporal scales based on the current state and task structure.

We evaluate the method across multiple environments with varying reward sparsity, spatial structure, and temporal constraints, including foraging, goal-reaching with hazards, and the Four Rooms. We further extend the setup to a continual learning scenario in which tasks change sequentially, requiring online adaptation of effective discounting.

Empirically, the proposed bio-inspired mixture-of-Q approach matches or outperforms individual fixed-discount experts across tasks. The results demonstrate that optimal temporal discounting is task-dependent and that adaptively combining multiple temporal horizons provides a flexible and robust alternative to manually selecting a single discount factor.

}

\begin{document}

\maketitle  

\begin{abstract}
Effective decision-making in complex environments requires balancing short- and long-term outcomes. In reinforcement learning (RL), this trade-off is controlled by a fixed discount factor, imposing a single exponentially discounted planning horizon. In contrast, biological agents adapt their temporal horizons to changing circumstances, suggesting that effective decision-making requires reasoning across multiple timescales. We propose a multi-horizon RL approach that adaptively combines temporal horizons, eliminating the need for manual discount-factor tuning. This makes the method well suited to continual learning, where tasks and environments evolve over time. Empirically, we demonstrate that our approach identifies effective discount factors in continual learning settings with three sequential task switches, suggesting that adaptive temporal discounting improves parameter efficiency and adaptability in both artificial and biological learning systems.


\end{abstract}

\section{Introduction}

A central challenge in sequential decision making is selecting the appropriate temporal horizon over which future outcomes should influence present actions. In reinforcement learning, this horizon is typically determined by a fixed discount factor ($\gamma$), implicitly assuming that a single timescale is appropriate throughout learning. Biological systems, however, appear to integrate predictions across multiple temporal horizons and flexibly adjust their reliance on them as environmental demands change. Consistent with this view, recent neuroscience studies suggest that distinct dopamine neurons and striatal circuits encode value over different discounting timescales \citep{masset2025multiscale,Sousa2025,mohebi2024dopamine}. These findings raise the possibility that adaptive, multi-timescale learning is a fundamental principle of biological decision making and motivate reinforcement learning algorithms that can dynamically adjust their temporal horizon to cope with delayed, sparse, and uncertain rewards.

In continual learning, task changes often require an agent to shift between short- and long-horizon decision-making. Some tasks emphasize immediate outcomes, such as avoiding imminent danger, whereas others require planning over extended horizons. This motivates our central research question: \textit{How can an agent dynamically adapt its planning horizon to different task demands, similar to the flexible decision-making observed in biological systems?}

To address this question, we first investigate how reward configuration, sparsity, and frequency influence the effective planning horizon in MiniGrid environments. We then propose a multi-horizon value estimation framework that adaptively combines value estimates across multiple timescales using learnable weighting coefficients. Our approach enables context-dependent temporal horizon selection and improves reward acquisition in environments with diverse temporal reward structures.

\section{Related Work}

Multi-timescale decision-making in reinforcement learning is closely tied to the discount factor $\gamma$, which determines the effective planning horizon. Performance is highly sensitive to this choice, and although meta-gradient methods \citep{hessel2019inductive} can adapt ($\gamma$) online, they still rely on a single exponentially discounted horizon. In contrast, behavioral and neuroscientific evidence suggests that biological agents learn and plan across multiple temporal scales with heterogeneous discounting \citep{laibson_golden_1997,vanderveldt_delay_2016,sozou1998hyperbolic}. 

Several works extend RL beyond a single discount factor \citep{fedus2019hyperbolic,Kurth-Nelson2009,redish_addiction_2004}. Hyperbolic discounting can be represented as a mixture of exponentially discounted value functions (with fixed weights) \citep{fedus2019hyperbolic}, while TD($\Delta$) decomposes value into multiple smaller temporal components to accelerate learning \citep{romoff2019timescales}. However, these methods do not provide a mechanism for adaptively selecting planning horizons in continually changing environments.

Our approach draws inspiration from mixture-of-experts (MoE) and value decomposition methods \citep{sunehag2018vdn}, which address the challenge of learning complex value functions by decomposing them into multiple simpler
components. Unlike QMIX \citep{rashid2018qmix}, which decomposes values across agents under a centralized training and decentralized execution (CTDE) framework, we combine horizon-specific critics, each learned with a different discount factor. A state-dependent gating network adaptively weights these critics, enabling flexible planning across multiple temporal horizons.

\section{Motivating Examples}

\subsection{Discount Factor and Optimal Policies}
In the following, we present two simple MDPs illustrating how varying $\gamma$ leads to different policies. 
Following the standard reinforcement learning formulation of Sutton and Barto~\cite{sutton2018reinforcement}, the return is defined as the discounted cumulative future reward: \(G_t = \sum_{k=0}^{\infty}\gamma^k R_{t+k+1}\), where \(\gamma \in [0,1]\) is the discount factor and \(R_{t+k+1}\) denotes the reward received at time step \(t+k+1\). The action-value function (Q-value) is defined as the expected return after taking action \(a\) in state \(s\) and subsequently following policy \(\pi\): \(Q_\pi(s,a)=\mathbb{E}_\pi[G_t \mid S_t=s,A_t=a]\). In deterministic environments, the transition dynamics and rewards are fixed, and therefore the expectation reduces to the deterministic return: \(Q(s,a)=G_t\).

\begin{figure}[htbp]
    \centering
    
    \begin{subfigure}[b]{0.49\textwidth}
    \centering
\includegraphics[width=0.5 \textwidth]{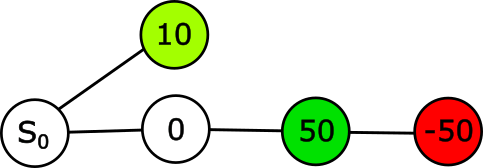}
    \caption{Deterministic hazard}
    \label{fig:threat}
    \end{subfigure}
    \hfill
    \begin{subfigure}[b]{0.49\textwidth}
    \centering
\includegraphics[width=0.65 \textwidth]{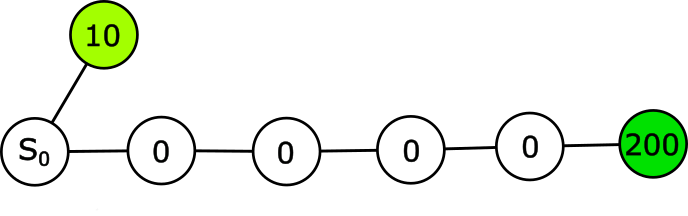}
    \caption{Trap-jackpot}
    \label{fig:trapjack}
    \end{subfigure}
    
    \caption{Fork Environments.}
    \label{fig:gamma_metrics}
\end{figure}

For the deterministic hazard fork in Figure~\ref{fig:threat}, Table~\ref{det_hazard_table} reports the Q-values of the left and right branches at the initial state. For the left branch, the agent receives an immediate reward of 10: \(G_{\text{Left}}=10\). For the right branch, the agent receives rewards of 0, 50, and -50 over subsequent time steps: \(G_{\text{Right}}=0+\gamma\cdot50+\gamma^2(-50)=50\gamma-50\gamma^2\). Thus, the greedy policy \(\pi(s)=\arg\max_a G(s,a)\) is dependent on the discount factor \(\gamma\): it selects the left branch for very small and very large values of \(\gamma\), and the right branch for intermediate values.

\begin{table}[htb!]
\centering
\scriptsize
\setlength{\tabcolsep}{3pt}
\renewcommand{\arraystretch}{0.75}

\begin{minipage}{0.48\textwidth}
\centering
\begin{tabular}{c c c c}
\hline
$\gamma$ & $Q_L$ & $Q_R$ & $\pi$ \\
\hline
0.10 & 10 & 4.50  & L \\
0.20 & 10 & 8.00  & L \\
0.30 & 10 & 10.50 & R \\
0.40 & 10 & 12.00 & R \\
0.50 & 10 & 12.50 & R \\
0.60 & 10 & 12.00 & R \\
0.70 & 10 & 10.50 & R \\
0.80 & 10 & 8.00  & L \\
0.90 & 10 & 4.50  & L \\
1.00 & 10 & 0.00  & L \\
\hline
\end{tabular}
\caption{Deterministic hazard fork.}
\label{det_hazard_table}
\end{minipage}
\hfill
\begin{minipage}{0.48\textwidth}
\centering
\begin{tabular}{c c c c}
\hline
$\gamma$ & $Q_L$ & $Q_R$ & $\pi$ \\
\hline
0.10 & 10 & 0.02   & L \\
0.20 & 10 & 0.32   & L \\
0.30 & 10 & 1.62   & L \\
0.40 & 10 & 5.12   & L \\
0.50 & 10 & 12.50  & R \\
0.60 & 10 & 25.92  & R \\
0.70 & 10 & 48.02  & R \\
0.80 & 10 & 81.92  & R \\
0.90 & 10 & 131.22 & R \\
1.00 & 10 & 200.00 & R \\
\hline
\end{tabular}
\caption{Trap--jackpot fork.}
\label{trap_jack_table}
\end{minipage}

\caption{Q-values and resulting greedy actions in the two MDPs.}
\end{table}

\textbf{Trap-jackpot fork} is illustrated in Figure~\ref{fig:trapjack}, with corresponding Q-values reported in Table~\ref{trap_jack_table}. This environment consists of a short-term rewarding but suboptimal left branch and a delayed high-reward “jackpot” on the right branch. We observe a clear dependence on the discount factor $\gamma$: for $\gamma < 0.5$ the agent prefers the left branch, while for $\gamma > 0.5$ it switches to the right branch, demonstrating how the effective planning horizon determines whether the agent prioritizes immediate rewards or delayed outcomes.

\textbf{Connection to Continual Learning}

Having shown that the optimal policy depends on $\gamma$, we consider two variations of the trap-jackpot MDP in Figure~\ref{fig:trapjack}. In the first, the jackpot is always available, while in the second, it disappears after a time limit. In both cases, the episode continues until all rewards are collected. The agent can either collect the smaller nearby reward first and then the distant jackpot, or prioritize the jackpot and return for the smaller reward. Without a time constraint, collecting the nearby reward first is more efficient in terms of reward per step, as it avoids unnecessary traversal. This favors a shorter planning horizon initially. However, when the jackpot is time-limited, the agent must prioritize reaching the distant reward before it disappears, requiring a longer planning horizon. The continual learning challenge is therefore to dynamically adapt the planning horizon $\gamma$ to arbitrate between these different reward structures and achieve efficient behavior across tasks.

\subsection{Reward Sparsity and Discounting}
We study the effect of discounting under different forms of reward sparsity in the MiniGrid foraging environment. In particular, we examine sparsity from two perspectives:

\begin{itemize}
    \item \textbf{Reward Reachability}: the spatial configuration and accessibility of rewards within the environment.
    \item 
    \textbf{Reward Frequency}: how often rewards are encountered by the agent.
\end{itemize}

\subsubsection{Reward Reachability}

\begin{figure}[htbp]
    \centering

    \begin{subfigure}[b]{0.12\textwidth}
        \centering
        \includegraphics[width=\linewidth]{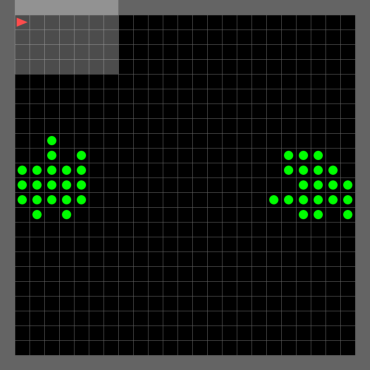}
    \end{subfigure}
    \hspace{0.01\textwidth}
    \begin{subfigure}[b]{0.12\textwidth}
        \centering
        \includegraphics[width=\linewidth]{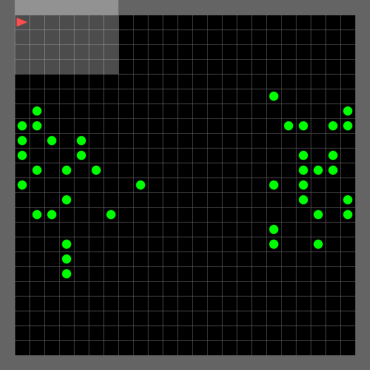}
    \end{subfigure}
    \hspace{0.01\textwidth}
    \begin{subfigure}[b]{0.12\textwidth}
        \centering
        \includegraphics[width=\linewidth]{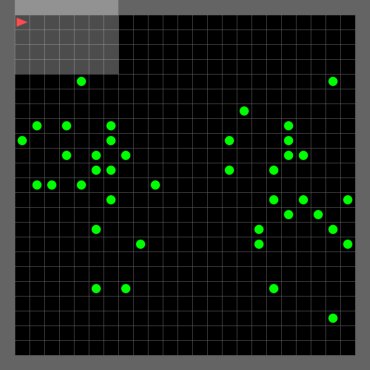}
    \end{subfigure}
    \hspace{0.01\textwidth}
    \begin{subfigure}[b]{0.12\textwidth}
        \centering
        \includegraphics[width=\linewidth]{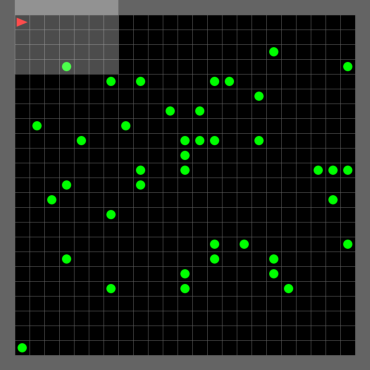}
    \end{subfigure}
    \hspace{0.01\textwidth}
    \begin{subfigure}[b]{0.12\textwidth}
        \centering
        \includegraphics[width=\linewidth]{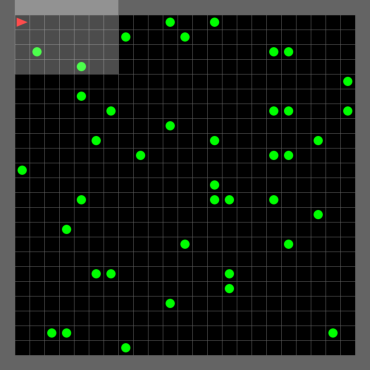}
    \end{subfigure}
    \caption{Reward configurations for different reward dispersions $(\sigma)$ ranging from 1 to 10.}
    \label{fig:reward_configs}
\end{figure}

To study the interaction between discounting and reward reachability, we generate reward locations by sampling from two Gaussian distributions. The Gaussian centers are placed at the midpoint of the grid along the vertical axis and at opposite ends along the horizontal axis. The agent’s objective is to collect all 40 rewards, distributed as 20 per Gaussian cluster, with each collected reward providing a unit reward of 1.
We vary the standard deviation ($\sigma$) of the Gaussian distributions from 1 to 10, creating environments with different levels of reward sparsity. As ($\sigma$) increases, rewards become more dispersed across the grid, increasing the distance and effort required for collection, as illustrated in Figure~\ref{fig:reward_configs}. We use tabular Expected SARSA($\lambda$) with 10 discount factors
$\gamma \in \{1,0.998,0.996,0.992,0.984,0.969,0.938,0.875,0.75,0.5\}$,
chosen to approximately double the effective horizon
$H_{\mathrm{eff}}=\frac{1}{1-\gamma}$ between consecutive values.
These correspond to horizons
$\{\infty,512,256,128,64,32,16,8,4,2\}$, respectively.

We evaluate performance using both the cumulative return and reward-per-step metrics at different stages of training as a function of gammas. Since the tasks are episodic, we use cumulative return as the primary performance metric. To further characterize the efficiency of different discount factors, we additionally analyze the average reward per step. The results after convergence (episode 9900), shown in Figure~\ref{fig:gamma_metrics}, indicate that the discount factor that maximizes performance shifts toward larger values as $\sigma$ increases. For $\sigma$ values between 1 and 5, the optimal discount factor is $\gamma=0.938$, corresponding to an effective horizon of approximately 16 steps. For $\sigma$ values between 6 and 10, the optimal discount factor increases to $\gamma = 0.969$, corresponding to an effective horizon of approximately 32 steps. These findings suggest that environments with more spatially dispersed rewards benefit from longer planning horizons and reduced discounting.
We observe the same effect across different step sizes, indicating that the observed pattern is not caused by the choice of step size. The step size is empirically tuned to 0.001 for the foraging task to achieve stable convergence and optimal performance. 

\begin{figure}[ht!]
    \centering
    
    \begin{subfigure}[b]{0.48\textwidth}
        \centering
        \includegraphics[width=\textwidth]{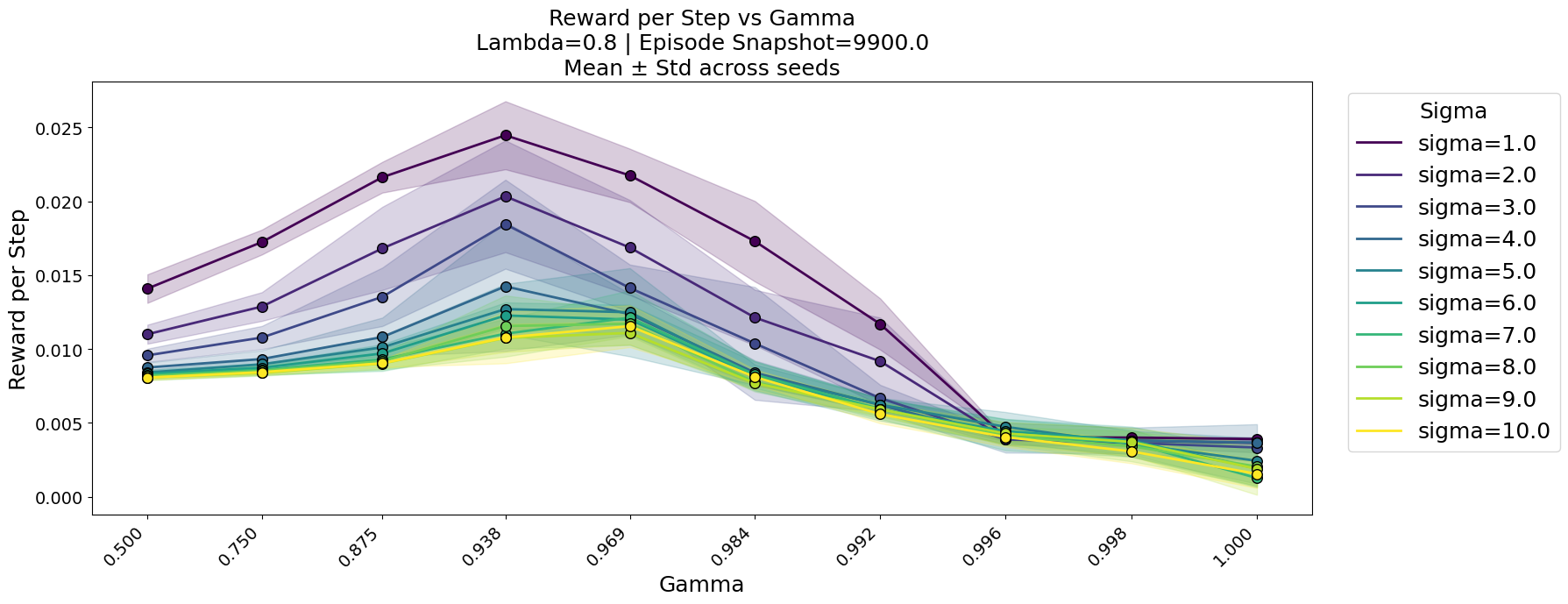}
        \caption{Reward per step vs. $\gamma$.}
        \label{fig:reward_per_step}
    \end{subfigure}
    \hspace{0.01\textwidth}
    \begin{subfigure}[b]{0.48\textwidth}
        \centering
        \includegraphics[width=\textwidth]{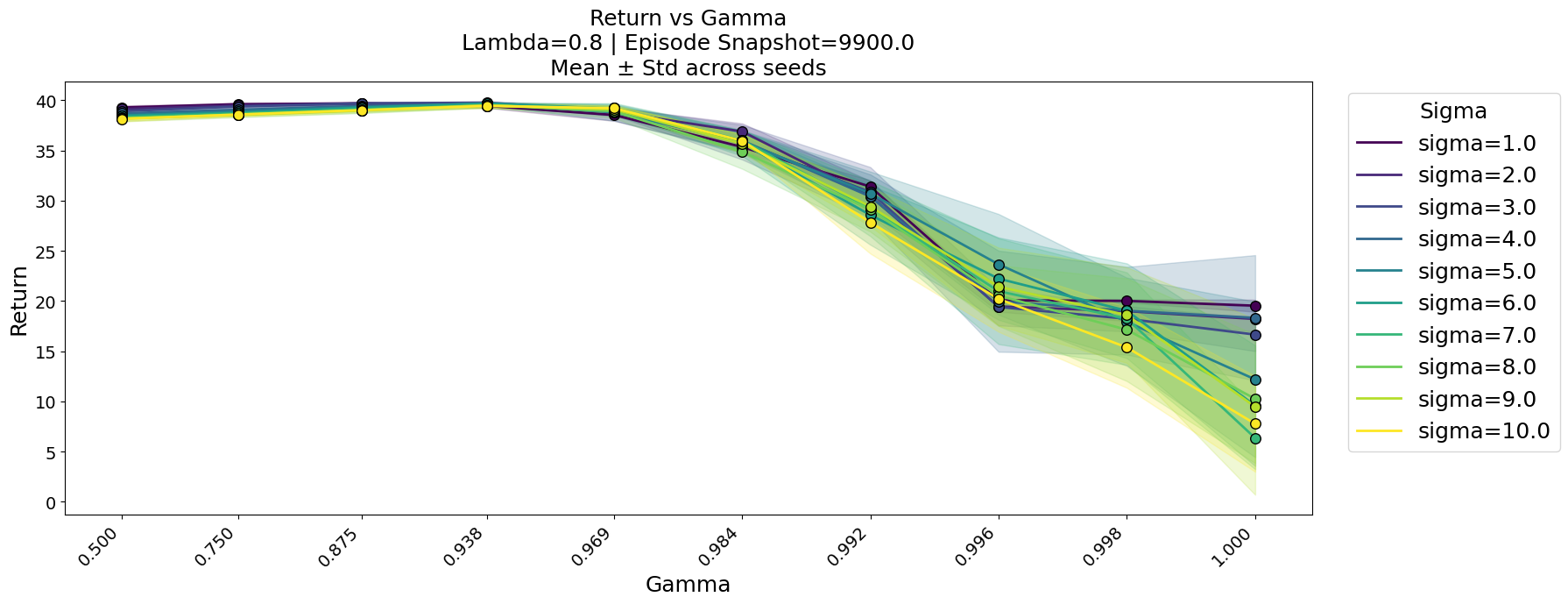}
        \caption{Return vs. $\gamma$.}
        \label{fig:return_vs_gamma}
    \end{subfigure}
    
    \caption{Performance, measured by return and reward per step, as a function of the discount factor $\gamma$ for different Gaussian reward distributions with $\sigma \in \{1,\dots,10\}$, averaged over 10 seeds. The trace parameter ($\lambda$) is fixed at 0.8, and the number of rewards per cluster is fixed at 20.}
    \label{fig:gamma_metrics}
\end{figure}

\subsubsection{Reward Frequency}

We also study the effect of reward frequency on the optimal discount factor, as shown in Fig.~\ref{fig:gamma_freq_rew} for ($\sigma$ = 10), Fig.~\ref{fig:gamma_freq_rew_sig8} for ($\sigma$ = 8), and in Appendix Fig.~\ref{fig:gamma_freq_rew_sig1} for ($\sigma$ = 1). We observe that as the number of rewards per cluster decreases from 20 to 2, larger discount factors yield better performance, as they mitigate reward sparsity by considering longer horizons. This effect becomes more pronounced for larger Gaussian standard deviations ($\sigma$), where rewards are more dispersed. Our environment design allows flexible testing across combinations of reward dispersion ($\sigma$) and the number of rewards per cluster, as shown in Appendix Figure~\ref{fig:sigma-num}. This enables systematic investigation of how environmental variations influence the optimal discount factor and other algorithmic parameters.

\begin{figure}[htbp!]
    \centering
    
    \begin{subfigure}[b]{0.49\textwidth}
        \centering
     \includegraphics[width=\textwidth]{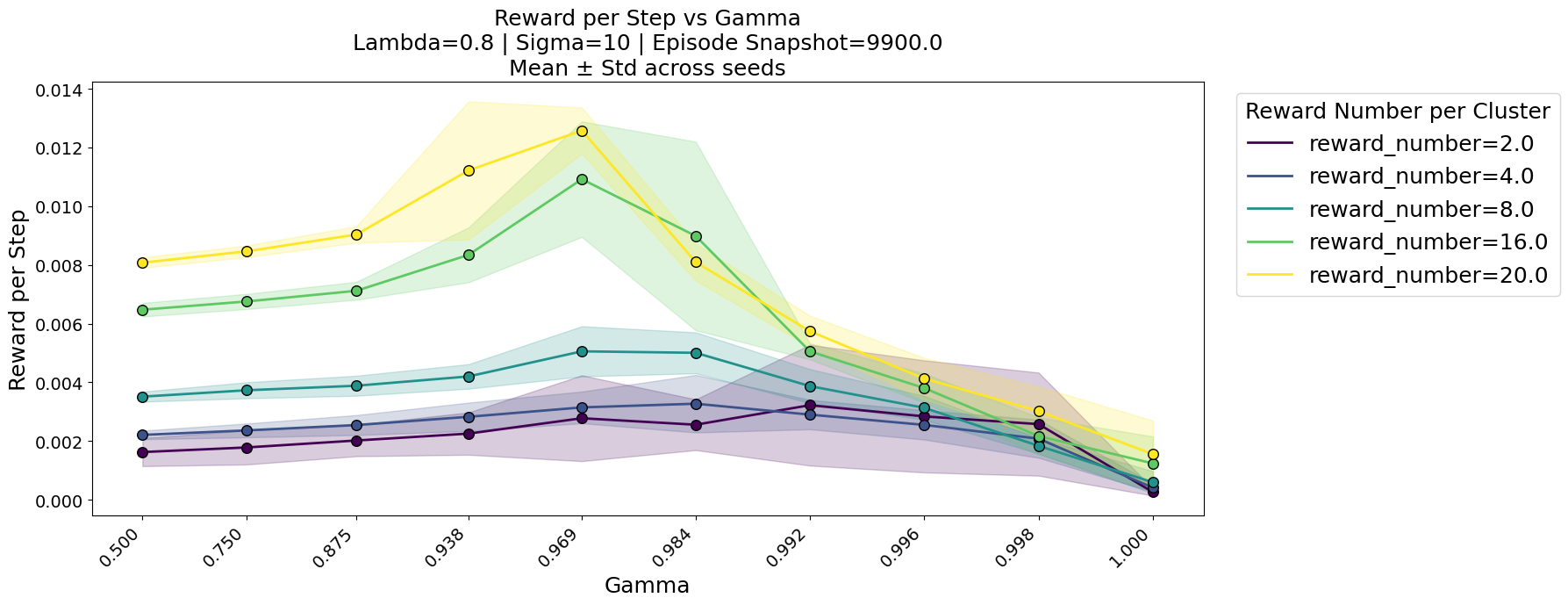}
        \caption{Reward per step vs. $\gamma$.}
        \label{fig:reward_per_step}
    \end{subfigure}
    \hfill
    \begin{subfigure}[b]{0.49\textwidth}
        \centering
        \includegraphics[width=\textwidth]{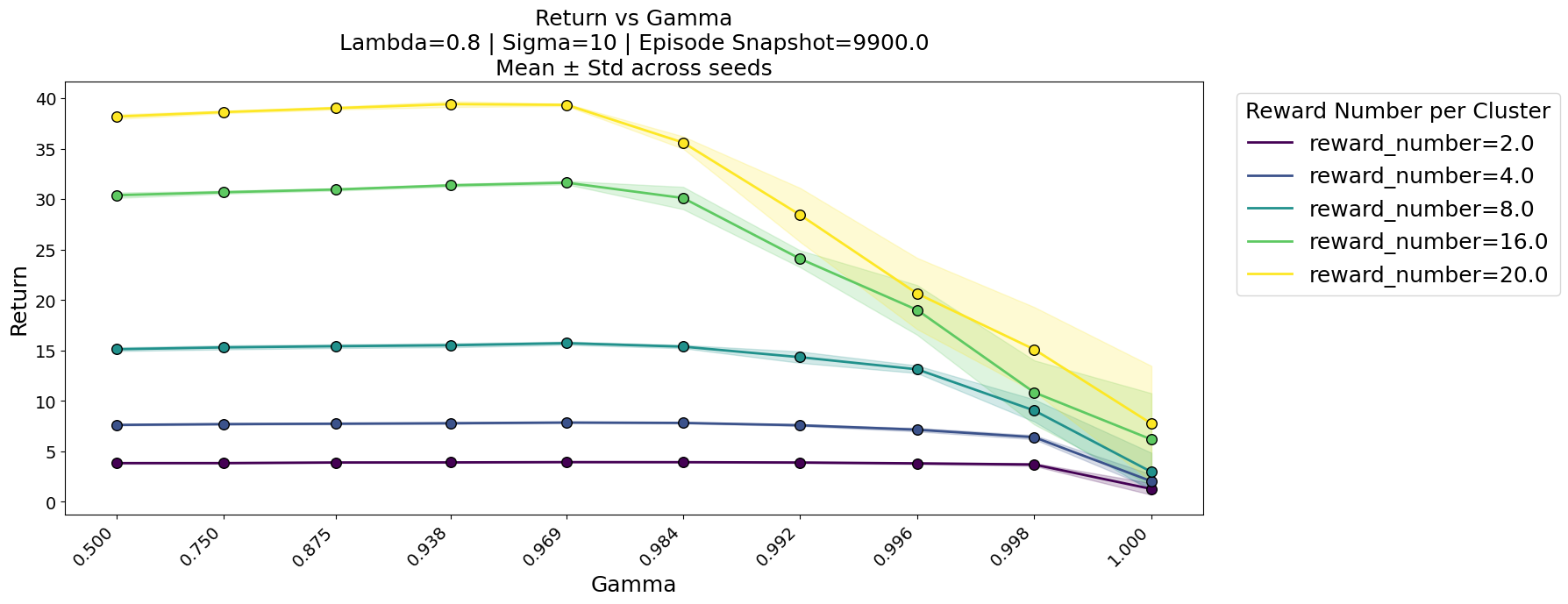}
        \caption{Return vs. $\gamma$.}
        \label{fig:return_vs_gamma}
    \end{subfigure}
    \caption{Return and reward per step, as a function of the discount factor ($\gamma$) for different numbers of rewards per cluster (${2, 4, 8, 16, 20}$), averaged over 5 seeds. The trace parameter ($\lambda$) is fixed at 0.8. Gaussian standard deviation ($\sigma$) is fixed at 10.}
    \label{fig:gamma_freq_rew}
\end{figure}

\begin{figure}[htbp!]
    \centering
    
    \begin{subfigure}[b]{0.49\textwidth}
        \centering
     \includegraphics[width=\textwidth]{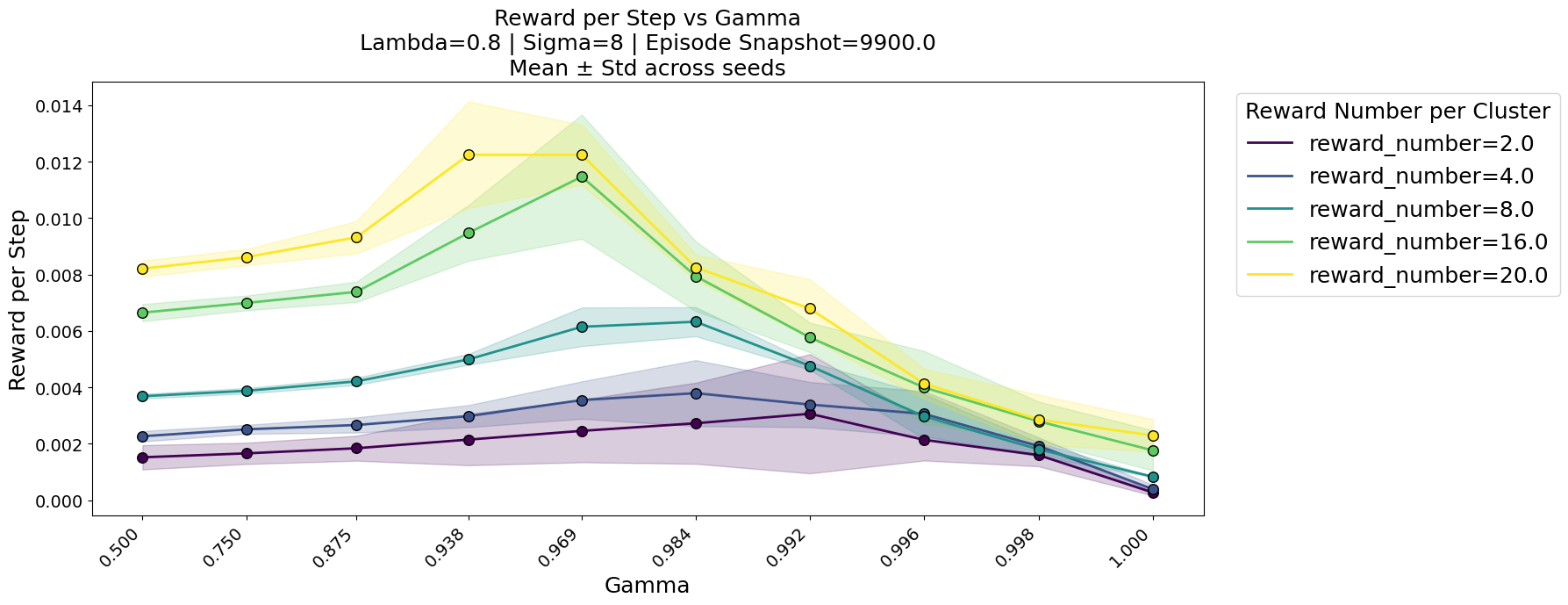}
        \caption{Reward per step vs. $\gamma$.}
        \label{fig:reward_per_step}
    \end{subfigure}
    \hfill
    \begin{subfigure}[b]{0.49\textwidth}
        \centering
        \includegraphics[width=\textwidth]{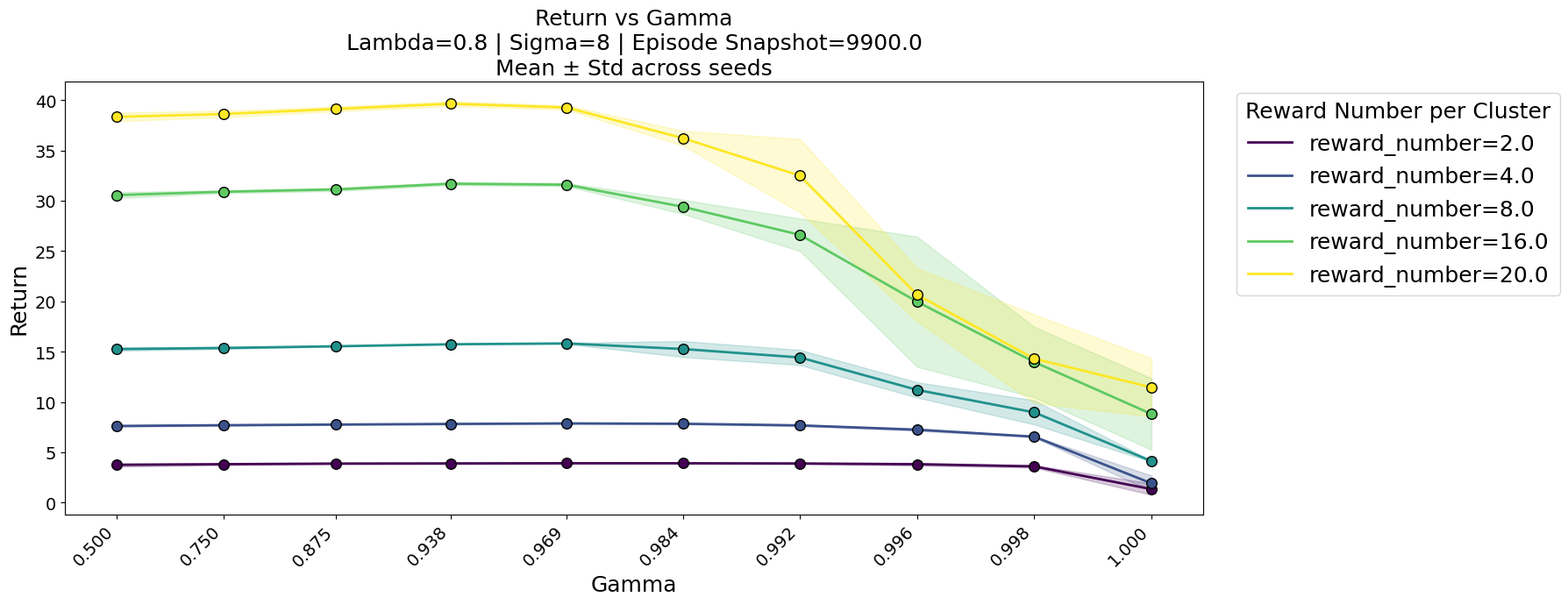}
        \caption{Return vs. $\gamma$.}
        \label{fig:return_vs_gamma}
    \end{subfigure}
    \caption{Return and reward per step, as a function of the discount factor ($\gamma$) for different numbers of rewards per cluster (${2, 4, 8, 16, 20}$), averaged over 5 seeds. The trace parameter ($\lambda$) is fixed at 0.8. Gaussian standard deviation ($\sigma$) is fixed at 8.}
    \label{fig:gamma_freq_rew_sig8}
\end{figure}
\vspace{-1em}
The dependency of the optimal discount factor on reward frequency and reachability highlights the need for adaptive discounting, particularly when switching between different environment configurations, such as the top-left and bottom-right grid settings in Figure~\ref{fig:reward_configs}.

\subsection{Task-Dependent Discounting}

We consider three tasks. Foraging, where the agent must collect 40 reward items distributed in a 25x25 grid within 5,000 steps. Goal Reaching, where the agent must reach a goal worth 40 points while avoiding 40 lava cells, each incurring a penalty of -0.1, within 2,500 steps. Four Rooms, where the environment contains a jackpot reward worth 36 points with a time limit of 2,116 steps and two local rewards per room (eight in total), each worth 0.5 points, all of which must be collected within a maximum of 6,250 steps.
 
\begin{figure}[ht!]
    \centering

    \begin{subfigure}[b]{0.1\textwidth}
        \centering
        \includegraphics[width=\textwidth]{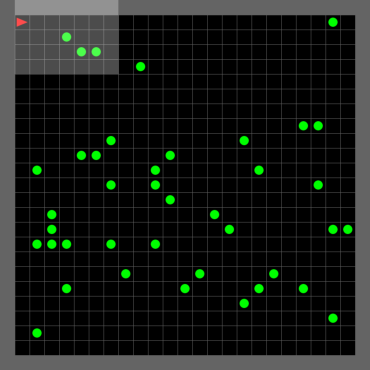}
        \caption*{Foraging}
        \label{fig:forg_}
    \end{subfigure}
    \hfill
    \begin{subfigure}[b]{0.44\textwidth}
        \centering
\includegraphics[width=\textwidth]{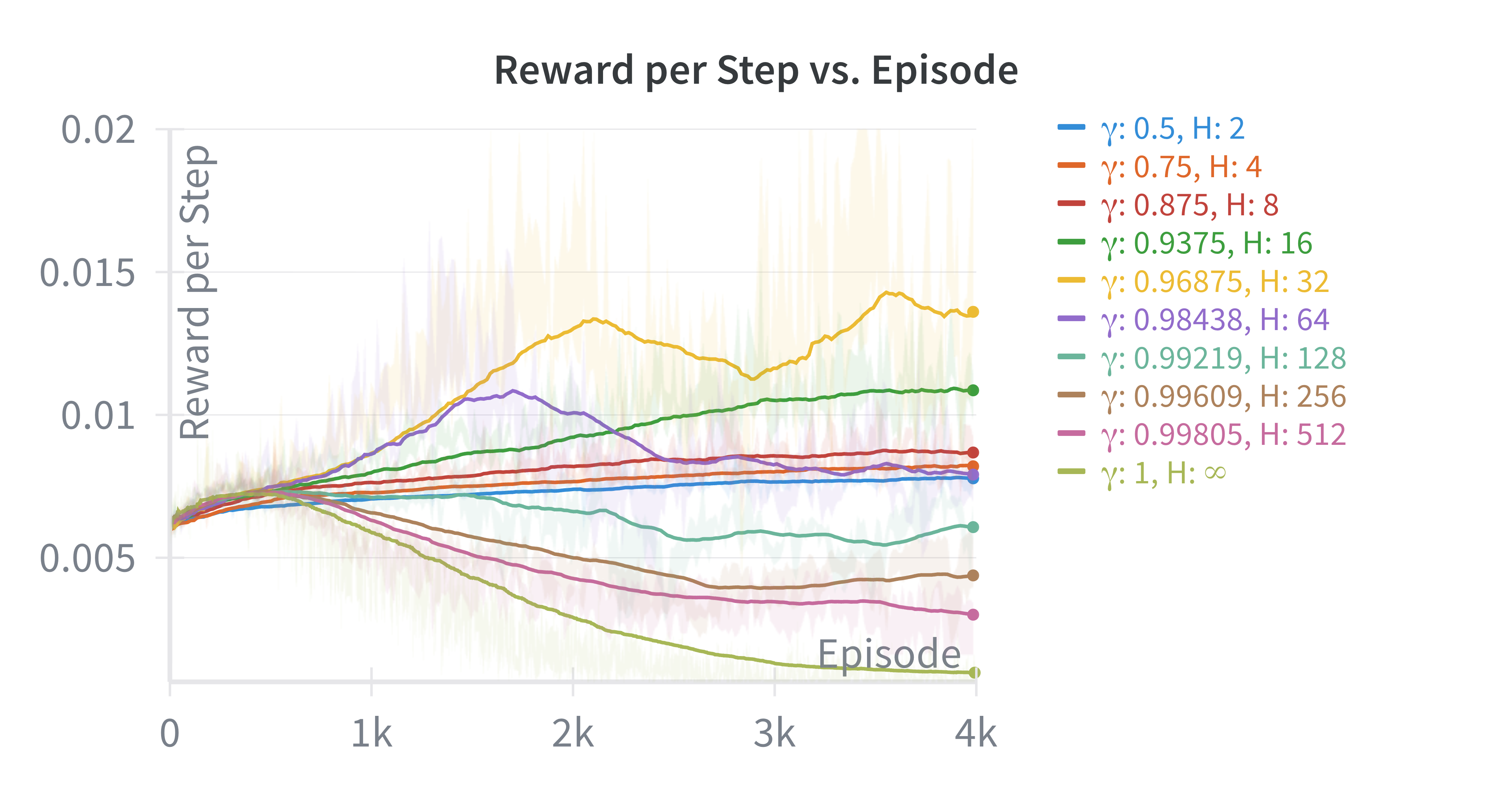}
        \caption{Reward per Step}
        \label{fig:forage_rewstep}
    \end{subfigure}
\hfill
    \begin{subfigure}[b]{0.44\textwidth}
        \centering
        \includegraphics[width=\textwidth]{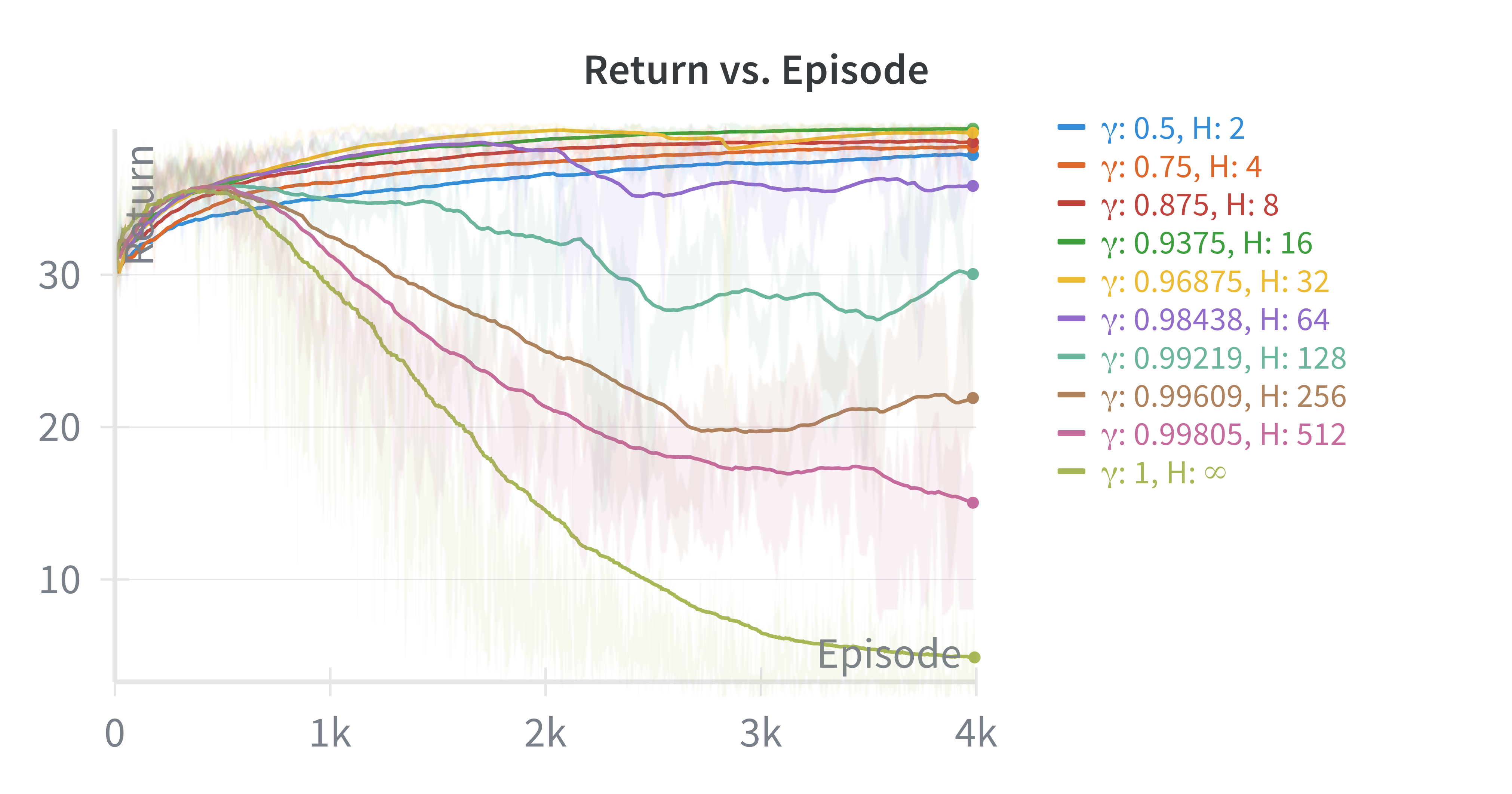}
        \caption{Return}
        \label{fig:forage_return}
    \end{subfigure}

    \caption{Performance metrics for different effective horizons in the foraging environment.}
    \label{fig:forage_}
\end{figure}

\begin{figure}[ht!]
    \begin{subfigure}[b]{0.1\textwidth}
        \centering \includegraphics[width=1\textwidth]{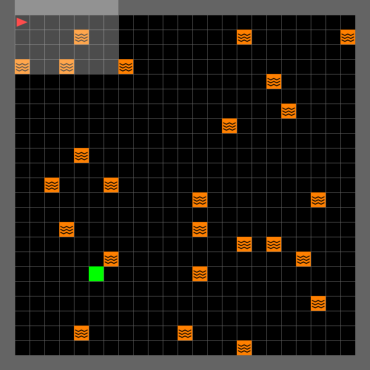}
        \caption*{Goal-lava}
        \label{fig:goal_env}
    \end{subfigure}
    \hfill
    \begin{subfigure}[b]{0.44\textwidth}
        \centering
    \includegraphics[width=1\textwidth]{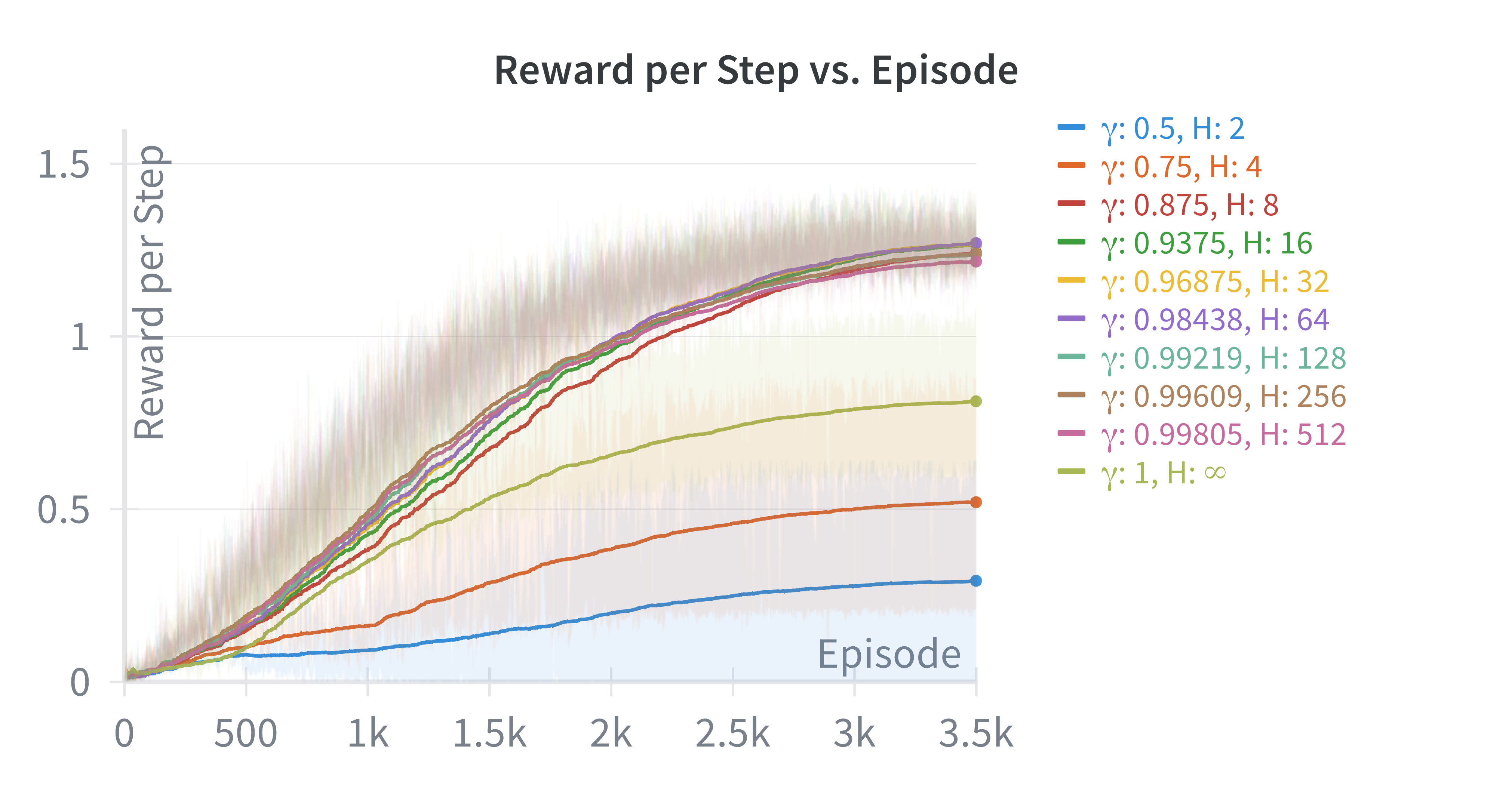}
        \caption{Reward per Step}
        \label{fig:goal_rewstep}
    \end{subfigure}
        \hfill
    \begin{subfigure}[b]{0.44\textwidth}
        \centering
\includegraphics[width=1\textwidth]{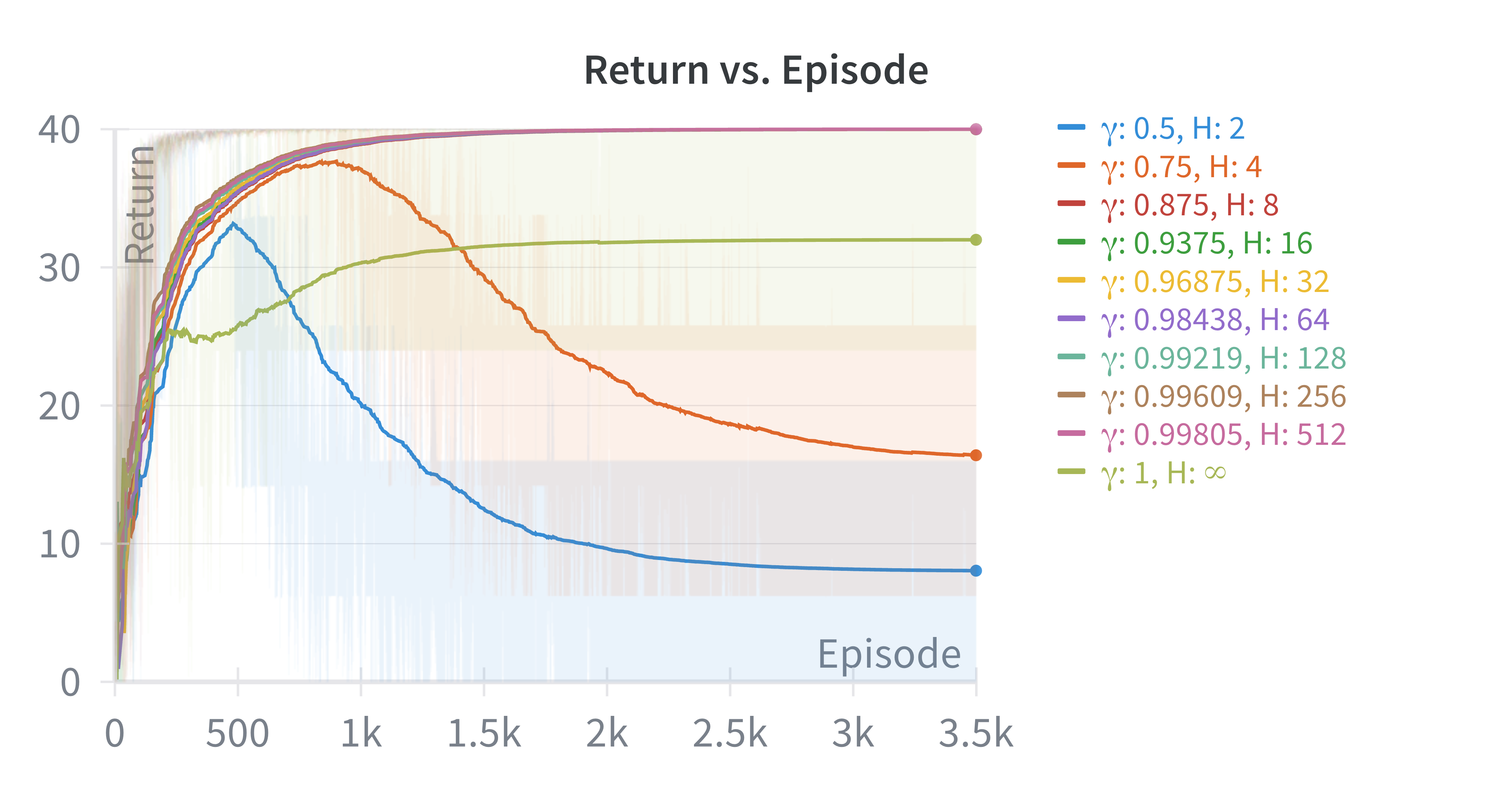}
        \caption{Return}
        \label{fig:goal_return}
    \end{subfigure}
    \caption{Performance metrics for different effective horizons in the goal reaching environment.}
    \label{fig:goal_}
\end{figure}

\begin{figure}[ht!]
    \begin{subfigure}[b]{0.1\textwidth}
        \centering \includegraphics[width=1\textwidth]{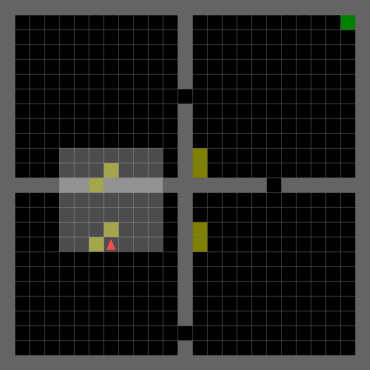}
        \caption*{Four-room}
        \label{fig:room_env}
    \end{subfigure}
    \hfill
    \begin{subfigure}[b]{0.44\textwidth}
        \centering
    \includegraphics[width=1\textwidth]{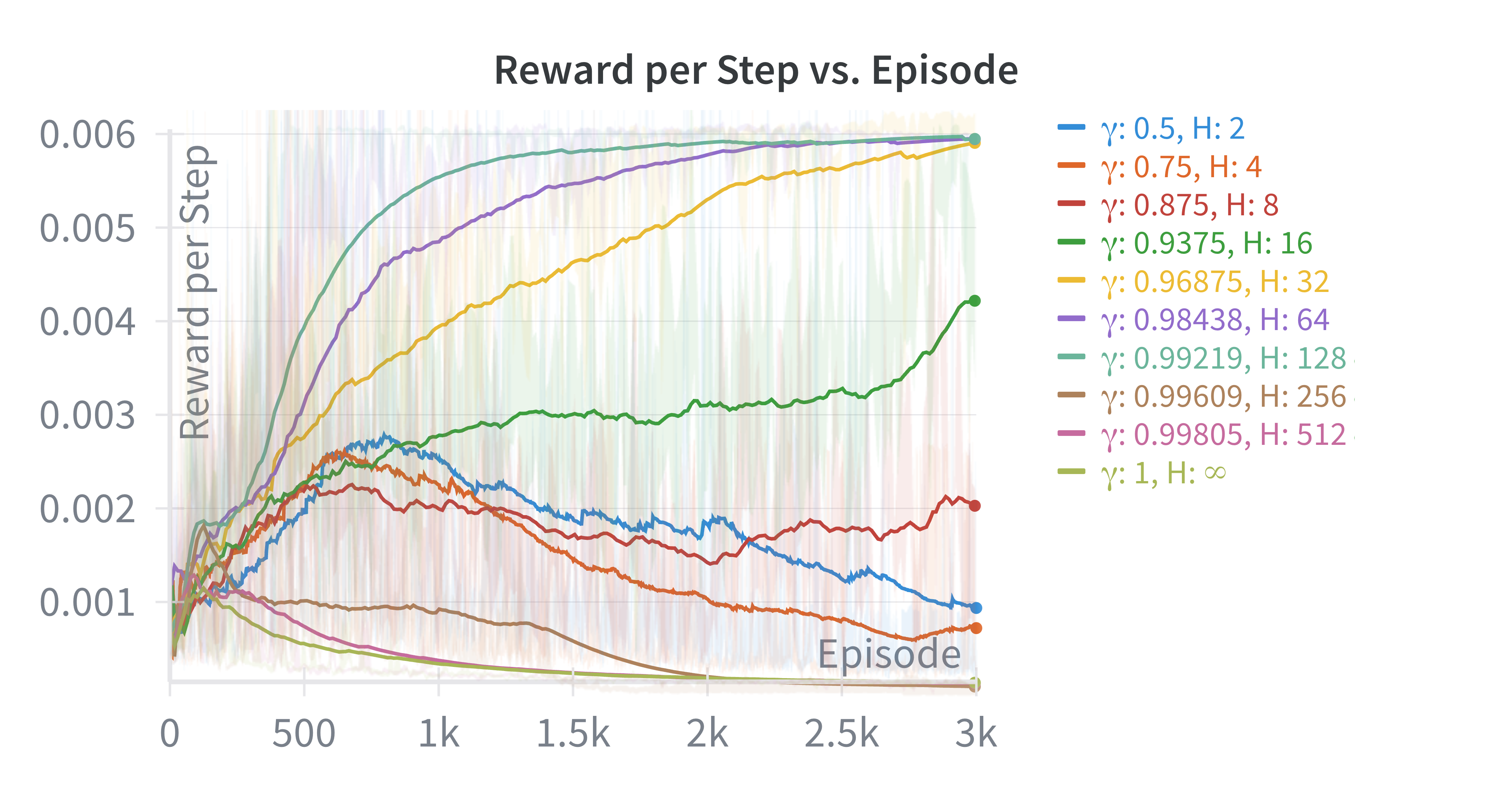}
        \caption{Reward per Step}
        \label{fig:room_rewstep}
    \end{subfigure}
        \hfill
    \begin{subfigure}[b]{0.44\textwidth}
        \centering
\includegraphics[width=1\textwidth]{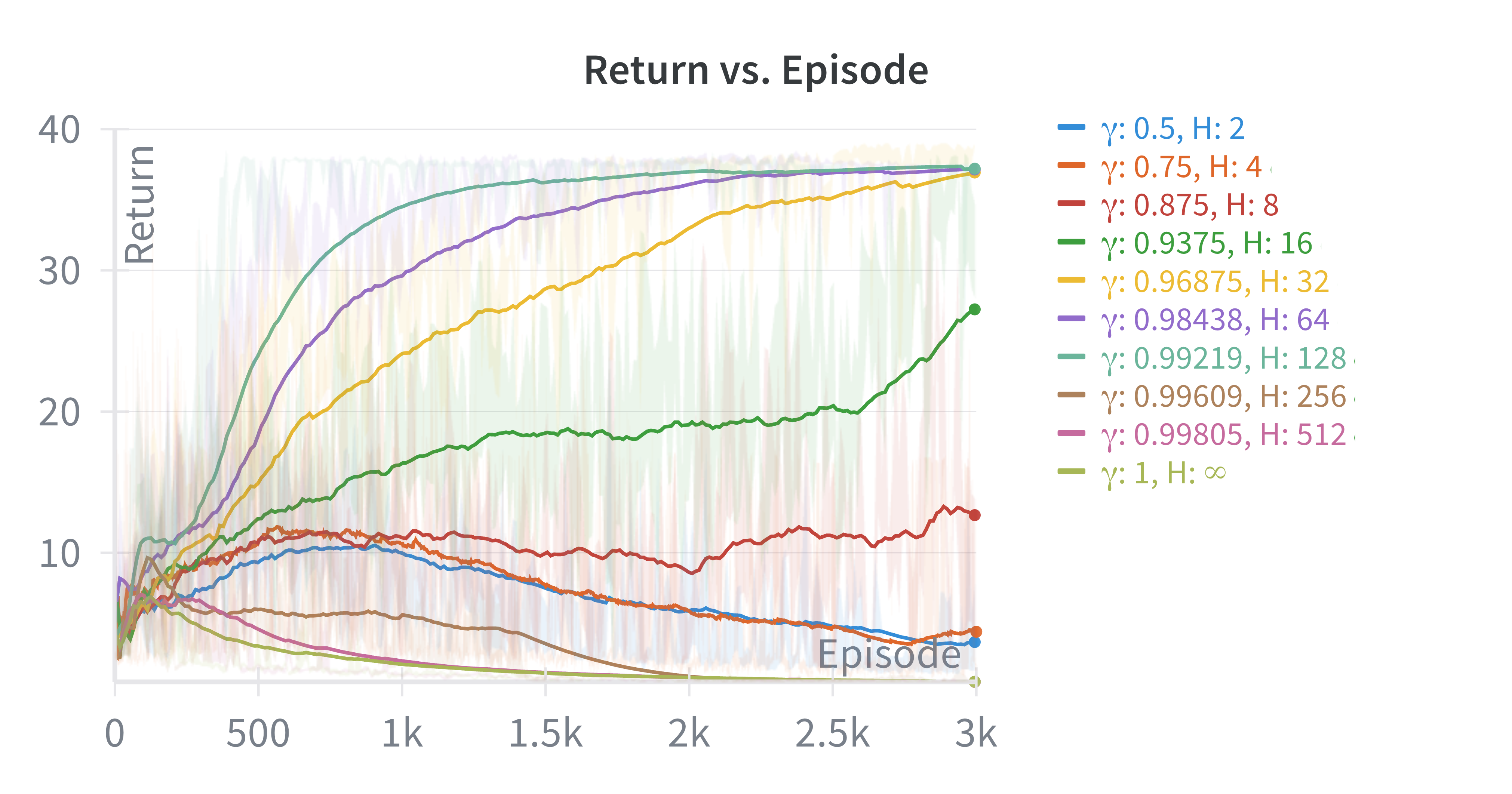}
        \caption{Return}
        \label{fig:room_return}
    \end{subfigure}
    \caption{Performance metrics for different effective horizons in the four rooms environment.}
    \label{fig:room_}
\end{figure}

As shown in Fig.~\ref{fig:forage_}, the best-performing discount factors in the Foraging task correspond to effective horizons of 16 and 32 steps. In Fig.~\ref{fig:goal_}, the optimal horizons for the Goal Reaching task lie between 32 and 512 steps. Finally, Fig.~\ref{fig:room_} shows in the Four Rooms task, the best performance is achieved with an effective horizon of 128 steps, while horizons in the range of 32–128 steps also perform competitively. All results are averaged over five seeds. 

The state representation used in these experiments is defined by the agent's (x,y) position in the grid and its orientation. The state space is finite and bounded by the grid size, and the action space is discrete and consists of three actions provided by MiniGrid: moving forward, rotating clockwise, and rotating counterclockwise. Unlike the partially observable setting commonly used in MiniGrid, the agent receives full grid observations in these experiments, providing complete access to the environment state. These findings highlight the importance of tuning the discount factor to the specific task and environment. This suggests that adaptive discounting can further improve performance in continual reinforcement learning settings involving task switches.

\section{Multi-Timescale Mixture of Q-Values}

We introduce a mixture-of-Q approach that dynamically adapts the effective
discount factor. We maintain multiple action-value functions
\(Q_i(s,a) \approx Q^\pi_{\gamma_i}(s,a)\) for a set of discount factors
\(\{\gamma_1,\dots,\gamma_K\}\). Each expert is learned
independently using tabular Expected SARSA($\lambda$) with
\(\lambda=0.8\).

For expert \(i\), the Expected SARSA($\lambda$) update is defined as
\(Q_i(s_t,a_t) \leftarrow Q_i(s_t,a_t)+\alpha\delta_t^{(i)}
E_t^{(i)}(s_t,a_t)\), where the TD error is
\(\delta_t^{(i)}=r_{t+1}+\gamma_i\sum_{a'}\pi(a'|s_{t+1})Q_i(s_{t+1},a')-Q_i(s_t,a_t)\),
and the eligibility trace is updated as
\(E_t^{(i)}(s,a)=\gamma_i\lambda E_{t-1}^{(i)}(s,a)+\mathbb{I}(s=s_t,a=a_t)\).

The policy is an \(\epsilon\)-greedy policy over the mixed value function
\(Q_{\mathrm{mix}}\)\footnotemark\footnotetext{Not to be confused with QMIX~\citep{rashid2018qmix}.}:
\(\pi(a|s)=\epsilon\text{-greedy}(Q_{\mathrm{mix}}(s,\cdot))\). The mixture weights are produced by a state-dependent gating network. Given a
state representation \(x=\phi(s)\), the network computes expert logits
\(z=Wx+b\), which are normalized using a softmax:
\(w_i(s)=\frac{\exp(z_i)}{\sum_k\exp(z_k)}\).
The mixed action-value function is then computed as
\(Q_{\mathrm{mix}}(s,a)=\sum_{i=1}^{K}w_i(s)Q_i(s,a)\). The mixture follows an undiscounted Expected SARSA update with
\(\gamma=1\), using the TD error
\(\delta=r_{t+1}+\sum_{a'}\pi(a'|s_{t+1})Q_{\mathrm{mix}}(s_{t+1},a')-Q_{\mathrm{mix}}(s_t,a_t)\).
The gating network parameters are optimized using the squared TD error loss
\(\mathcal{L}=\frac{1}{2}\delta^2\). The resulting semi-gradient updates for
the gating parameters are:
\begin{equation}
\begin{aligned}
W &\leftarrow W+
\alpha\delta
\left[w\odot(Q-Q_{\mathrm{mix}})\right]x^\top, \\
b &\leftarrow b+
\alpha\delta
\left[w\odot(Q-Q_{\mathrm{mix}})\right].
\end{aligned}
\end{equation}
We update the mixer weights using these equations every 50 steps based on trajectories sampled from a replay buffer. The full derivation of the gating gradients is provided in Appendix~\ref{app:derivation}.

\subsection{Experimental Results and Analysis}

\begin{figure}[htbp!]
    \centering

    \begin{subfigure}[t]{0.40\textwidth}
        \centering
        \includegraphics[width=\textwidth]{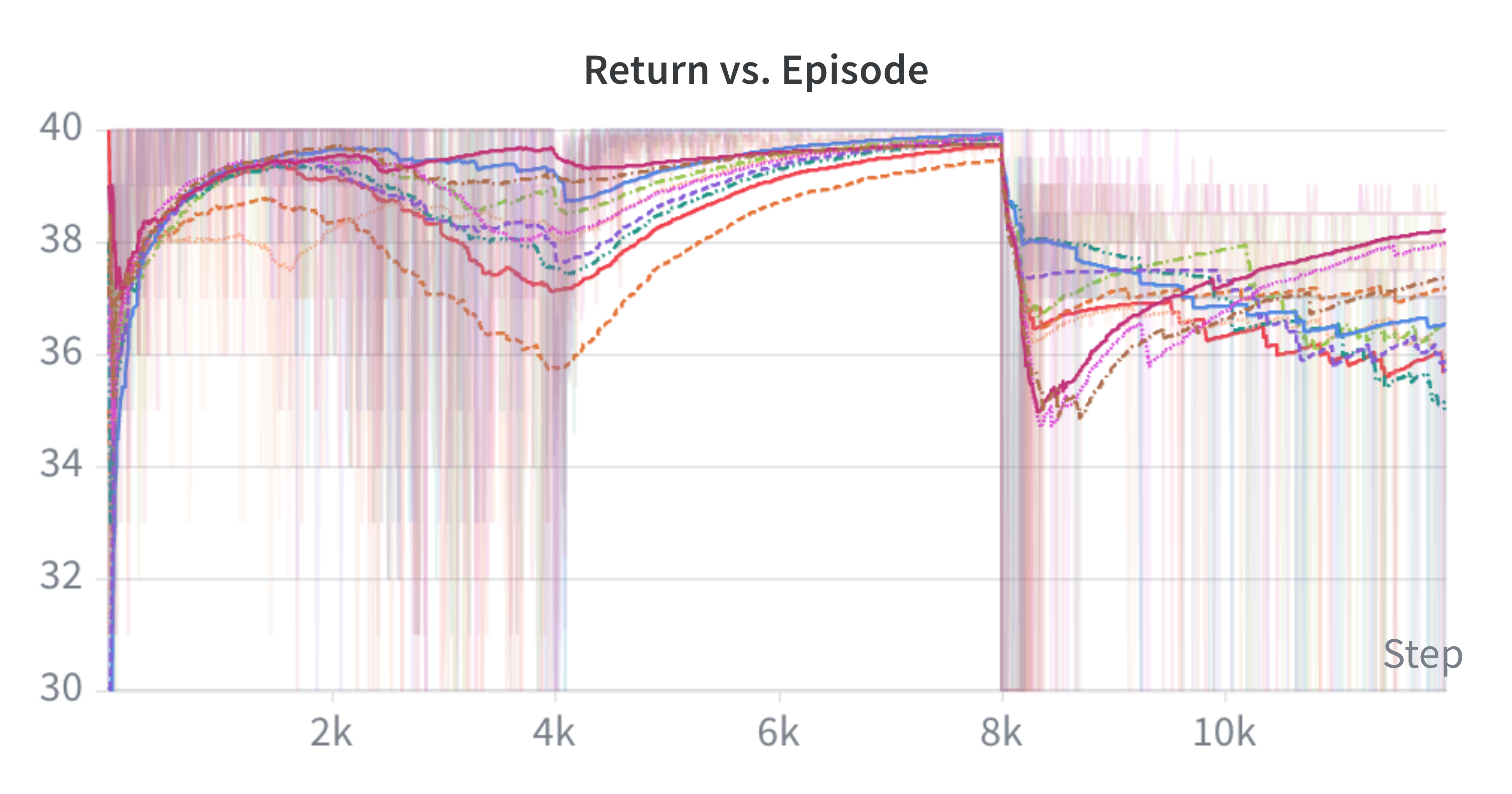}
        \caption{Episode return under continual task switching.}
        \label{fig:crlreturns}
    \end{subfigure}
    \hspace{0.02\textwidth}
    \begin{subfigure}[t]{0.40\textwidth}
        \centering
        \includegraphics[width=\textwidth]{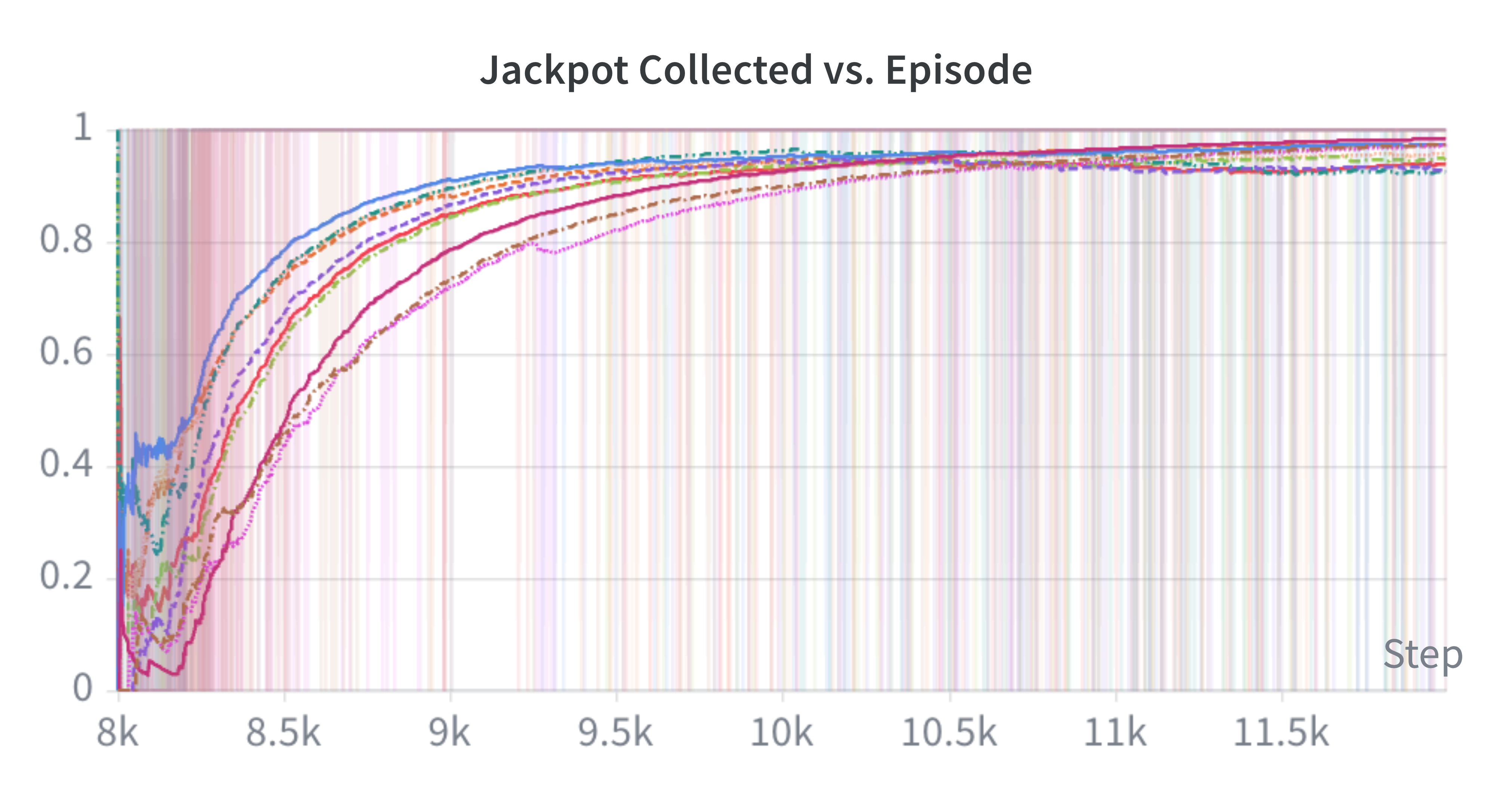}
        \caption{Time-constraint jackpot collection status (third environment).}
        \label{jackpot_}
    \end{subfigure}

    \vspace{0.2em}

    \begin{subfigure}[t]{0.40\textwidth}
        \centering
        \includegraphics[width=\textwidth]{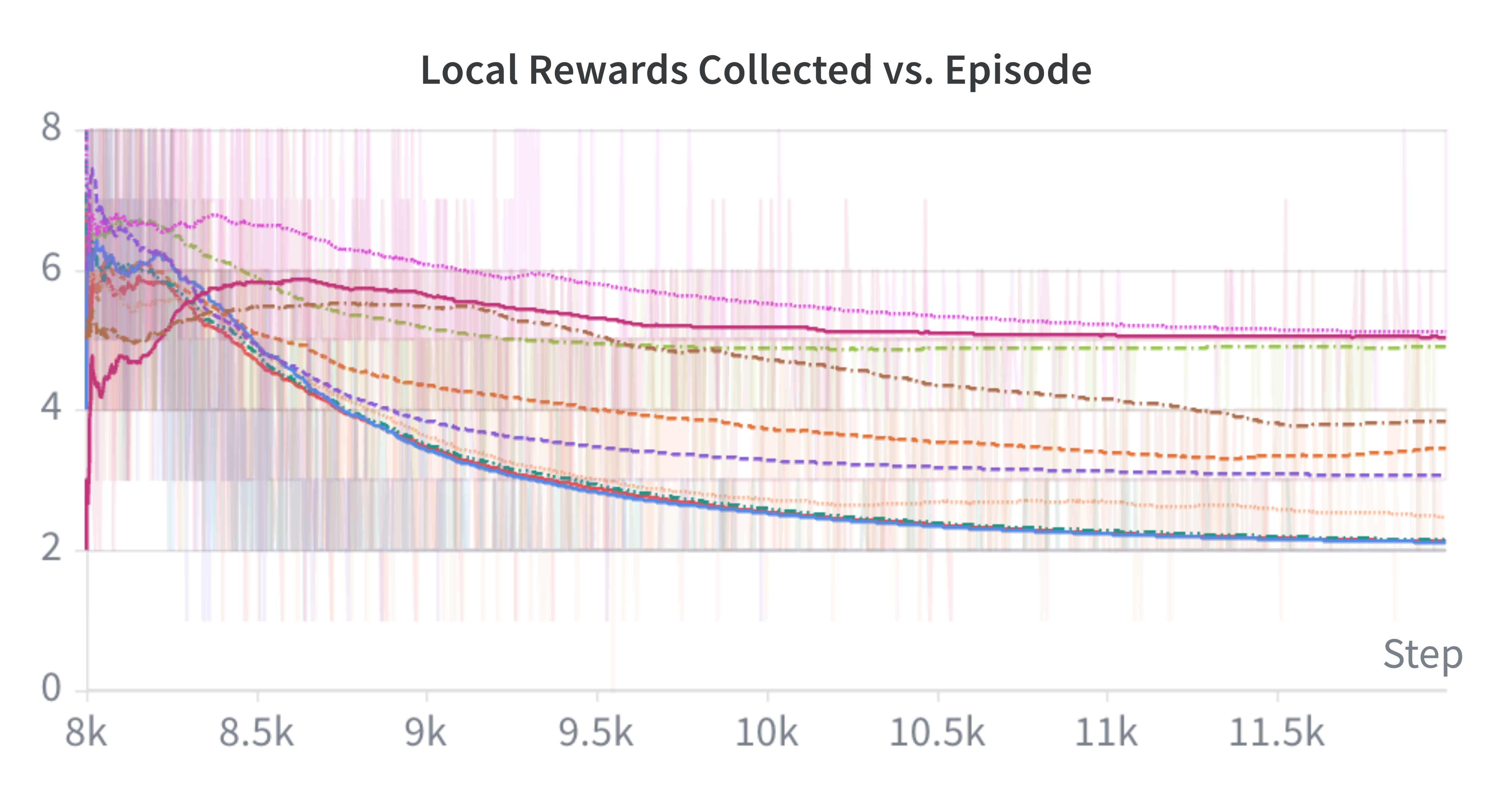}
        \caption{Local rewards collected after the jackpot (out of 8, third environment).}
        \label{local_}
    \end{subfigure}
    \hspace{0.02\textwidth}
    \begin{subfigure}[t]{0.40\textwidth}
        \centering
        \includegraphics[width=\textwidth]{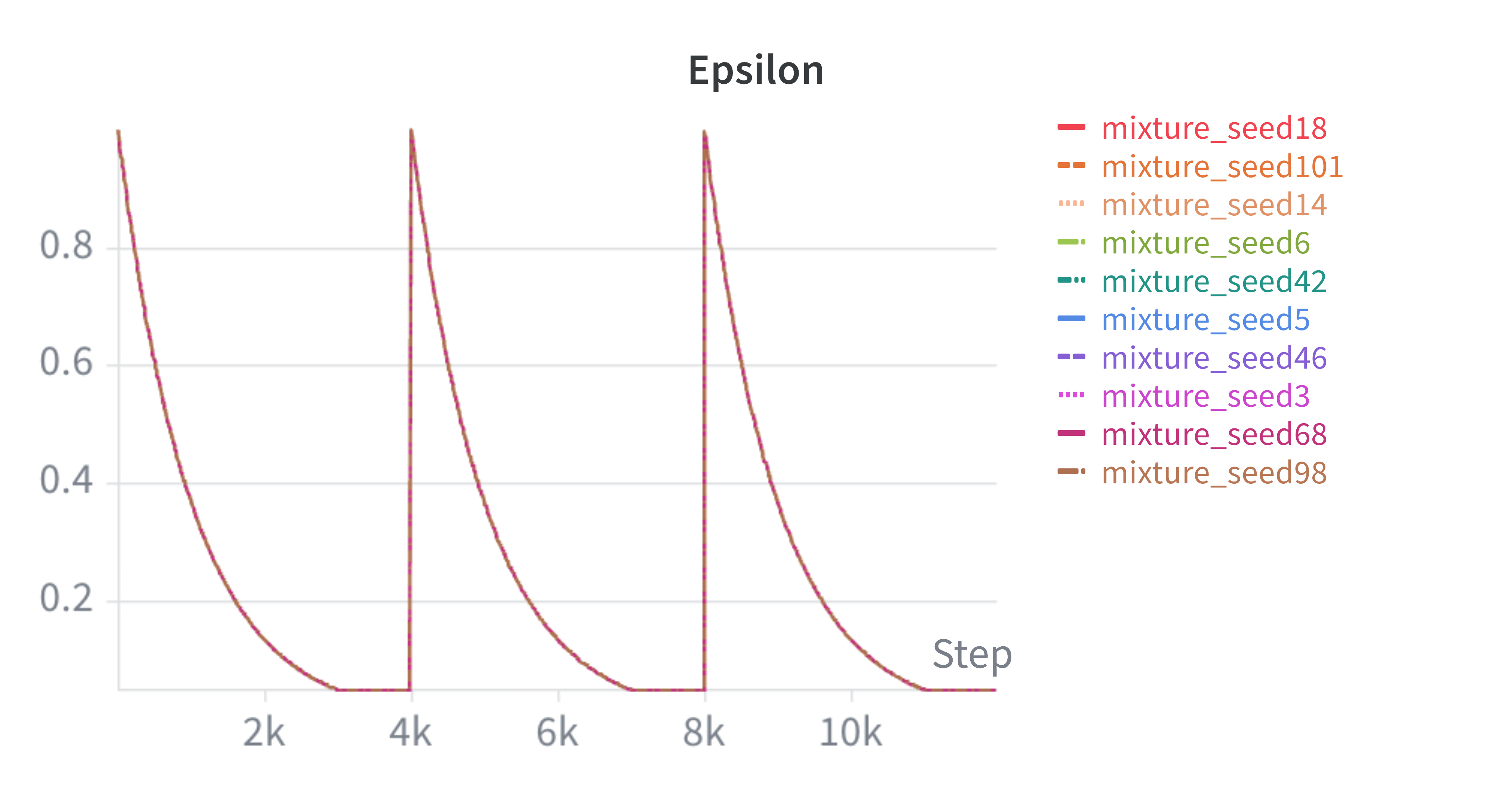}
        \caption{Epsilon exploration decay and reset between tasks.}
    \end{subfigure}

    \caption{Learning performance during continual task transitions.}
    \label{fig:CRL}
\end{figure}

We evaluate our approach in a continual learning setting, where the task transitions sequentially through the three environments shown in Appendix Figure ~\ref{fig:continual}. Our method consistently adapts to each task switch while maintaining near-optimal performance. We transition to a new task after 4k episodes, which was empirically selected as a sufficient duration for each task to converge. At each task transition, we reset the exploration rate ($\epsilon$) and learning rate ($\alpha$). The exploration rate is initialized to 1 and decayed to 0.05 with a decay rate of 0.999. The initial learning rate is task-dependent: 0.001 for the foraging task, 0.1 for the goal-reaching, and 0.01 for the four rooms, with these values empirically selected for each environment. The learning rate is then decayed throughout each task to one-tenth of its initial value with a decay rate of 0.9995. The results across 10 seeds are shown in Fig.~\ref{fig:CRL}.

\begin{figure}[htbp!]
    \centering

    \begin{subfigure}[b]{0.31\textwidth}
        \centering
        \includegraphics[width=\textwidth]{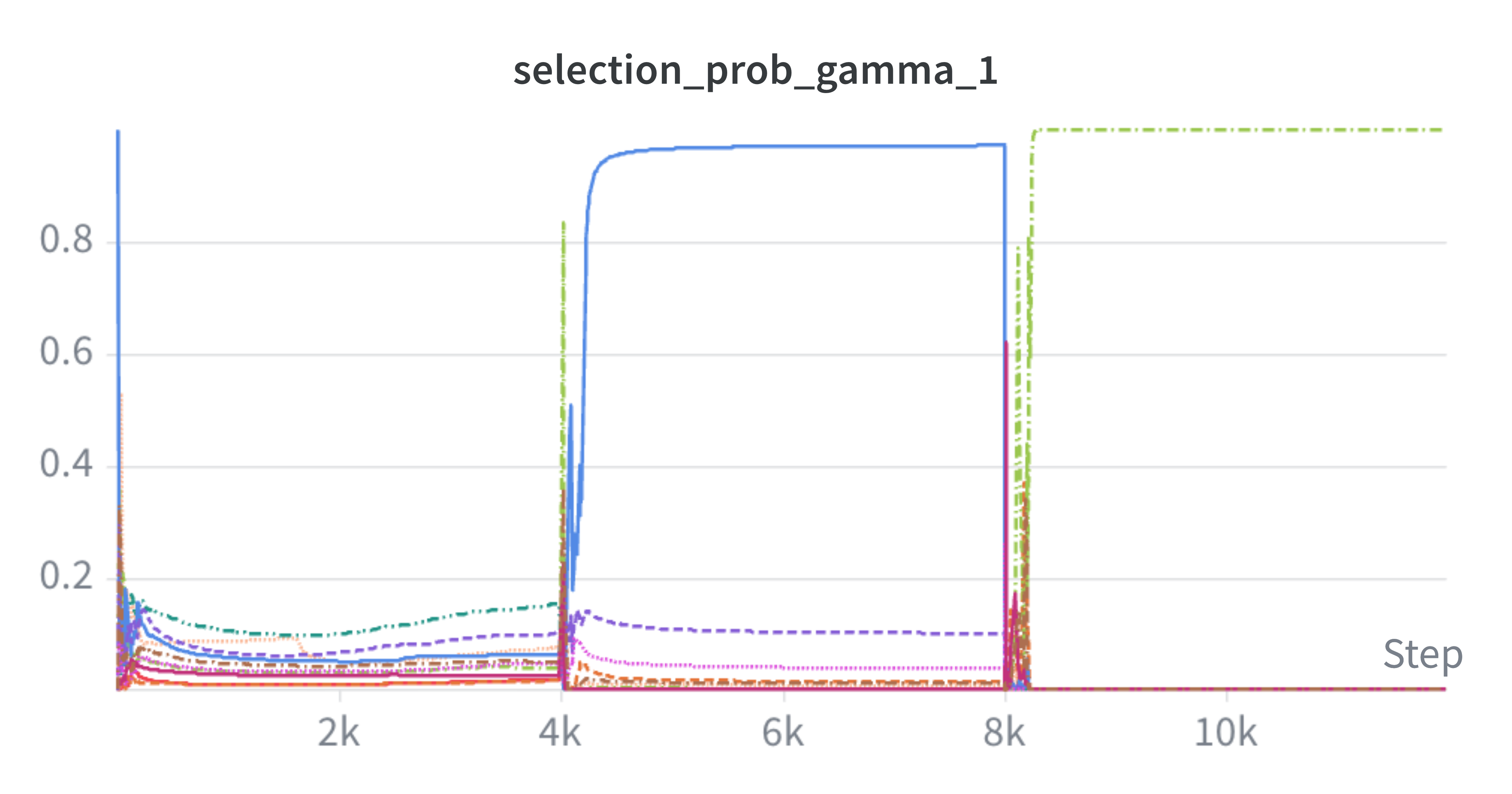}
    \end{subfigure}
    \hfill
    \begin{subfigure}[b]{0.31\textwidth}
        \centering
        \includegraphics[width=\textwidth]{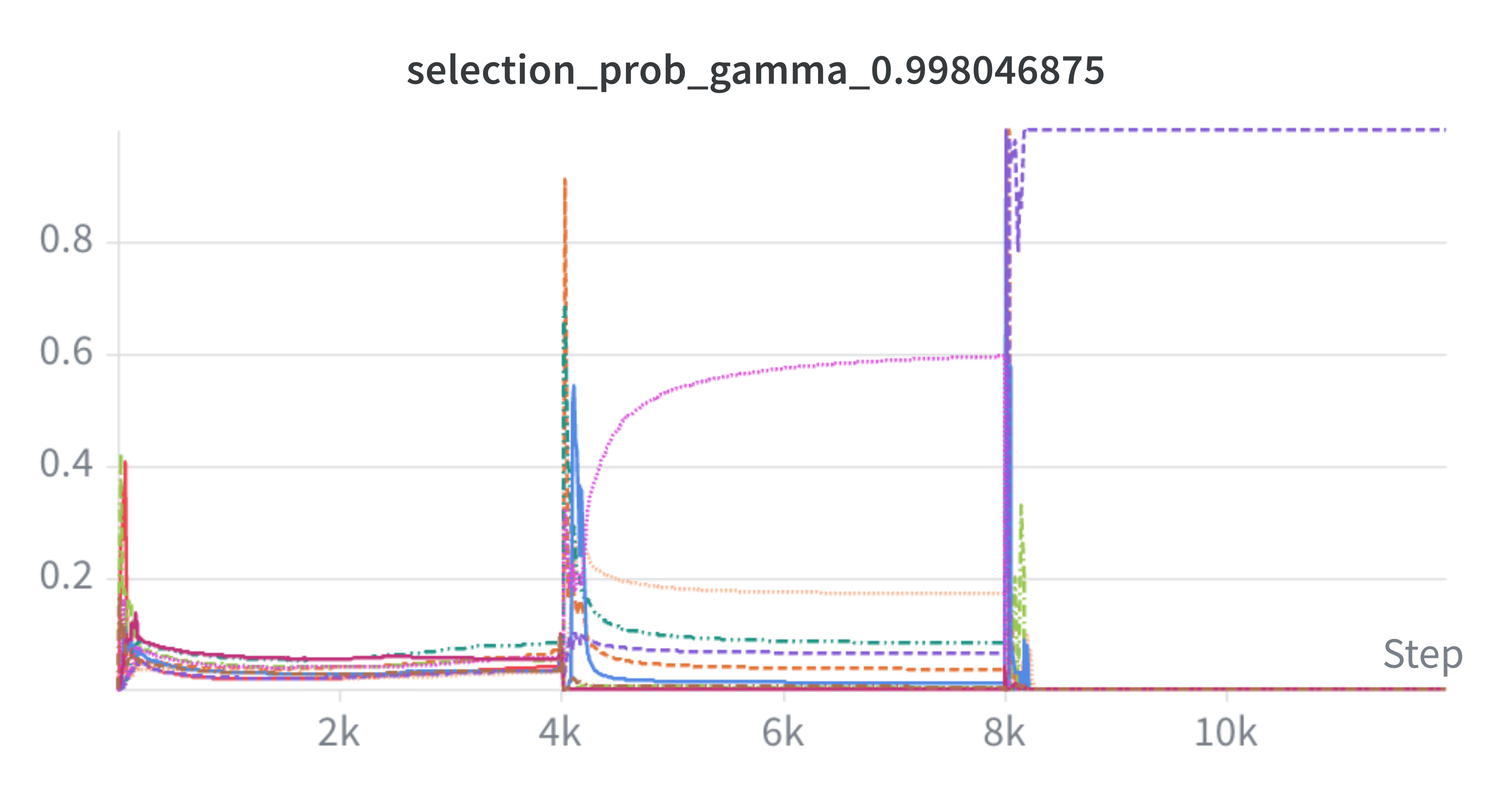}
    \end{subfigure}
    \hfill
    \begin{subfigure}[b]{0.31\textwidth}
        \centering
        \includegraphics[width=\textwidth]{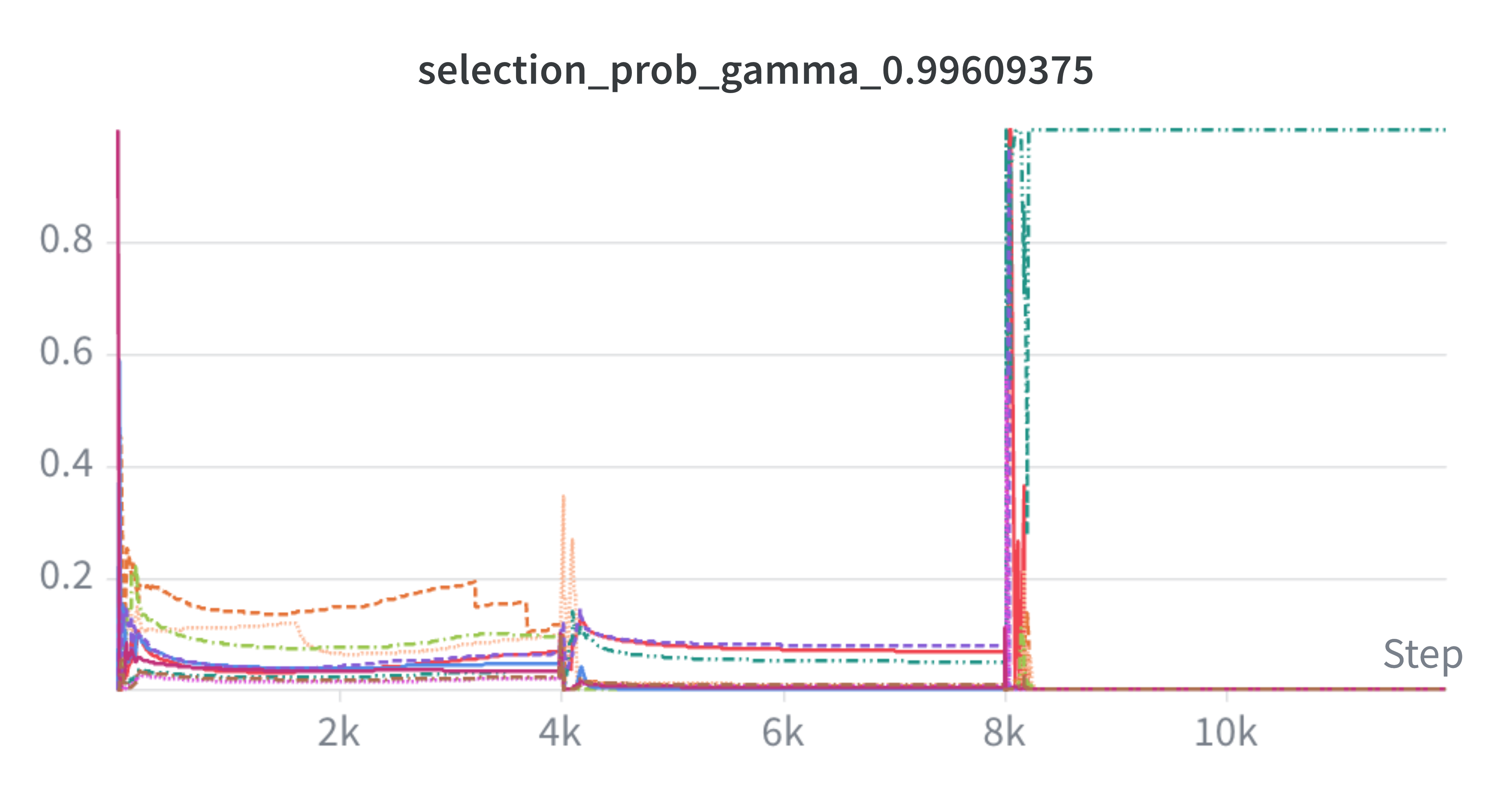}
    \end{subfigure}

    \vspace{0.2em}

    \begin{subfigure}[b]{0.31\textwidth}
        \centering
        \includegraphics[width=\textwidth]{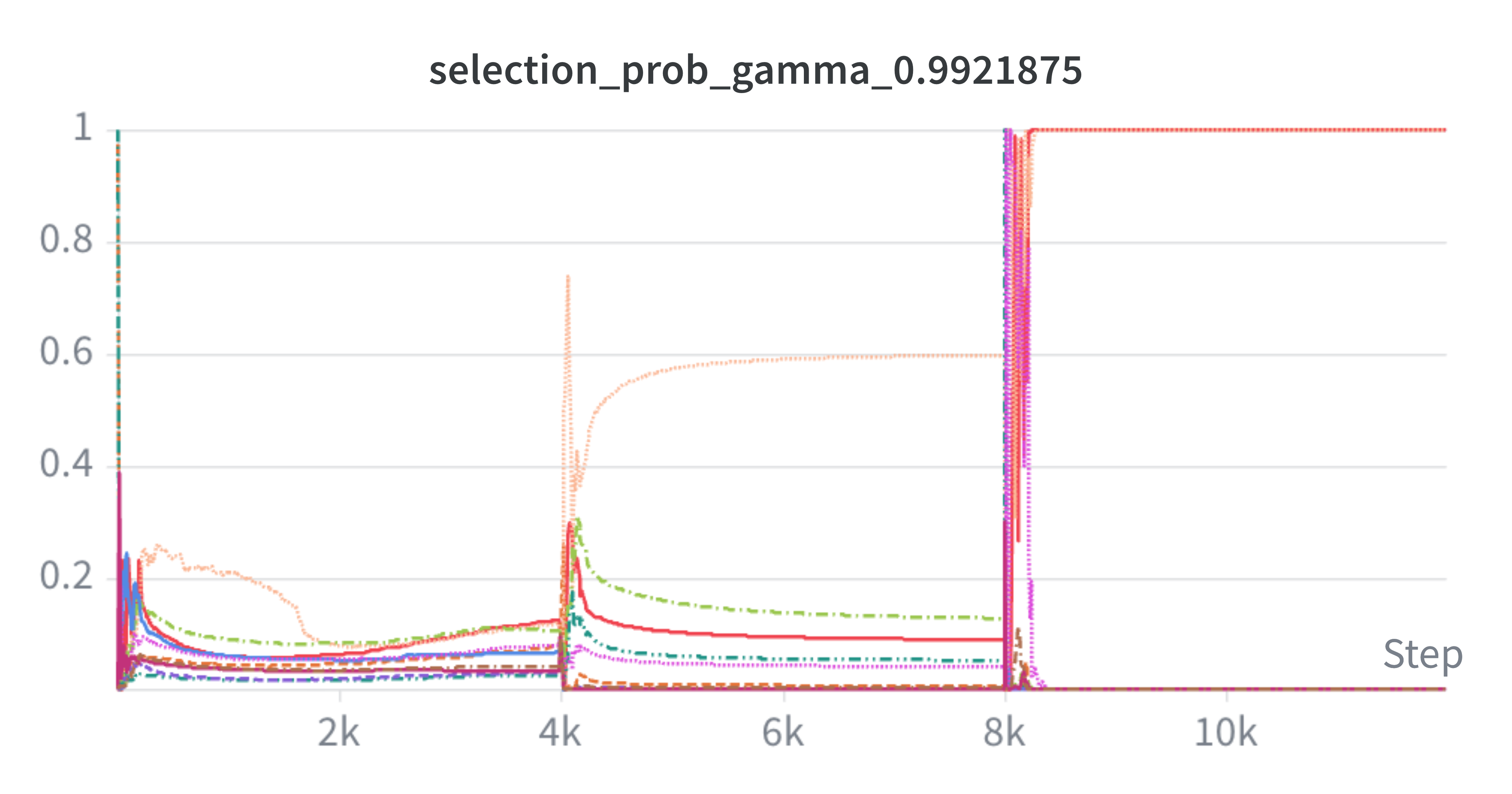}
    \end{subfigure}
    \hfill
    \begin{subfigure}[b]{0.31\textwidth}
        \centering
        \includegraphics[width=\textwidth]{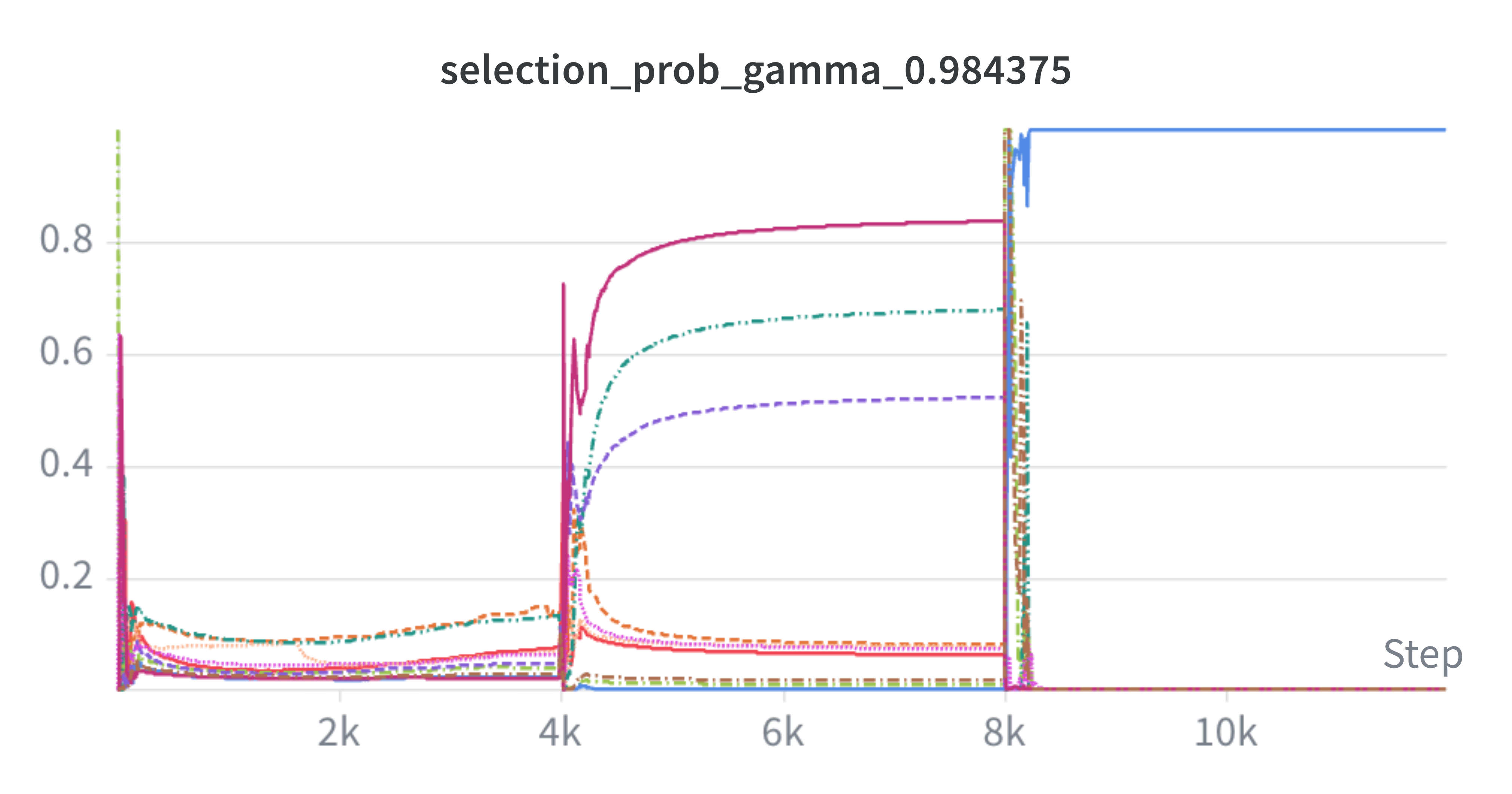}
    \end{subfigure}
    \hfill
    \begin{subfigure}[b]{0.31\textwidth}
        \centering
        \includegraphics[width=\textwidth]{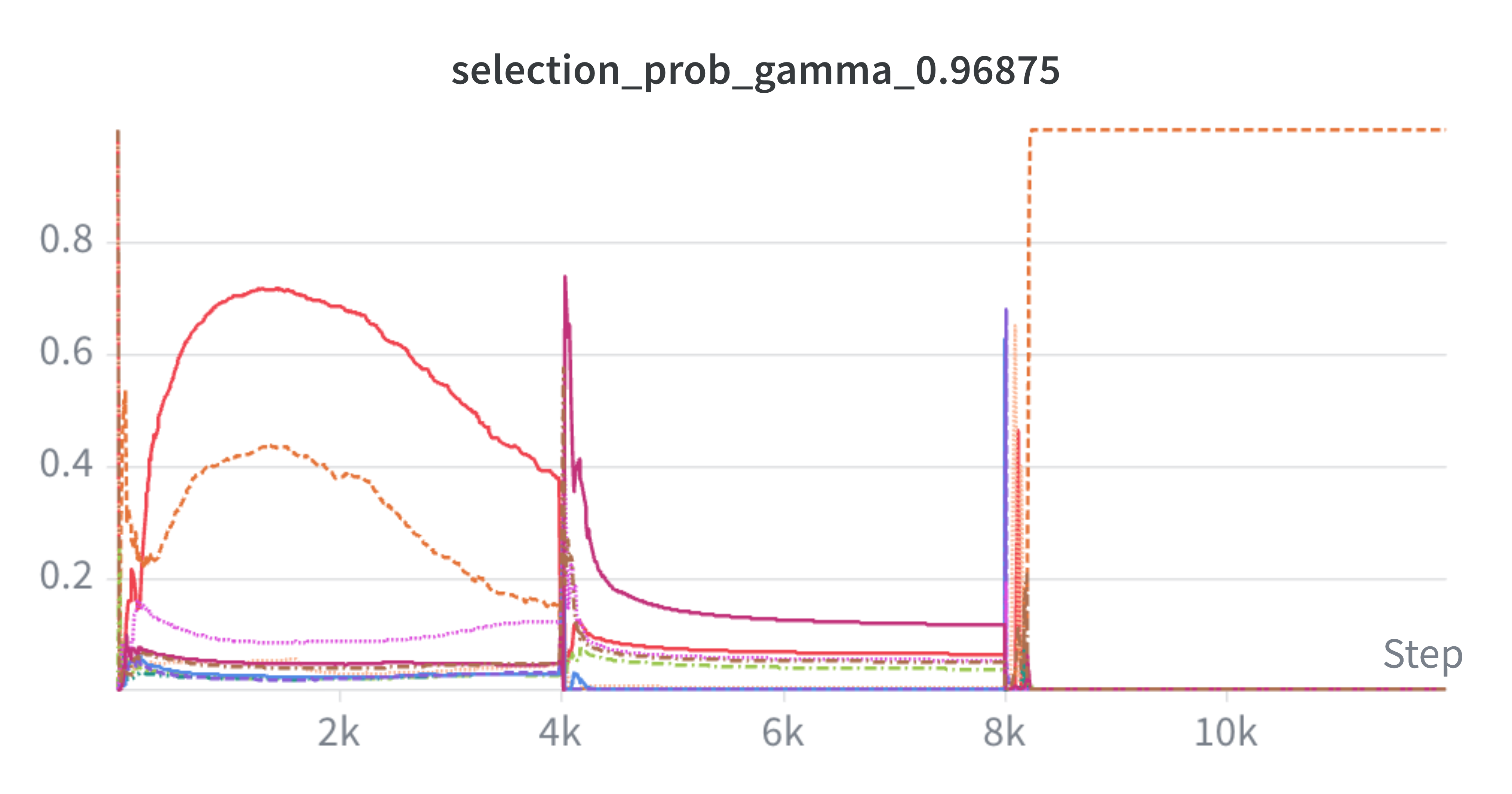}
    \end{subfigure}

    \vspace{0.2em}

    \begin{subfigure}[b]{0.31\textwidth}
        \centering
        \includegraphics[width=\textwidth]{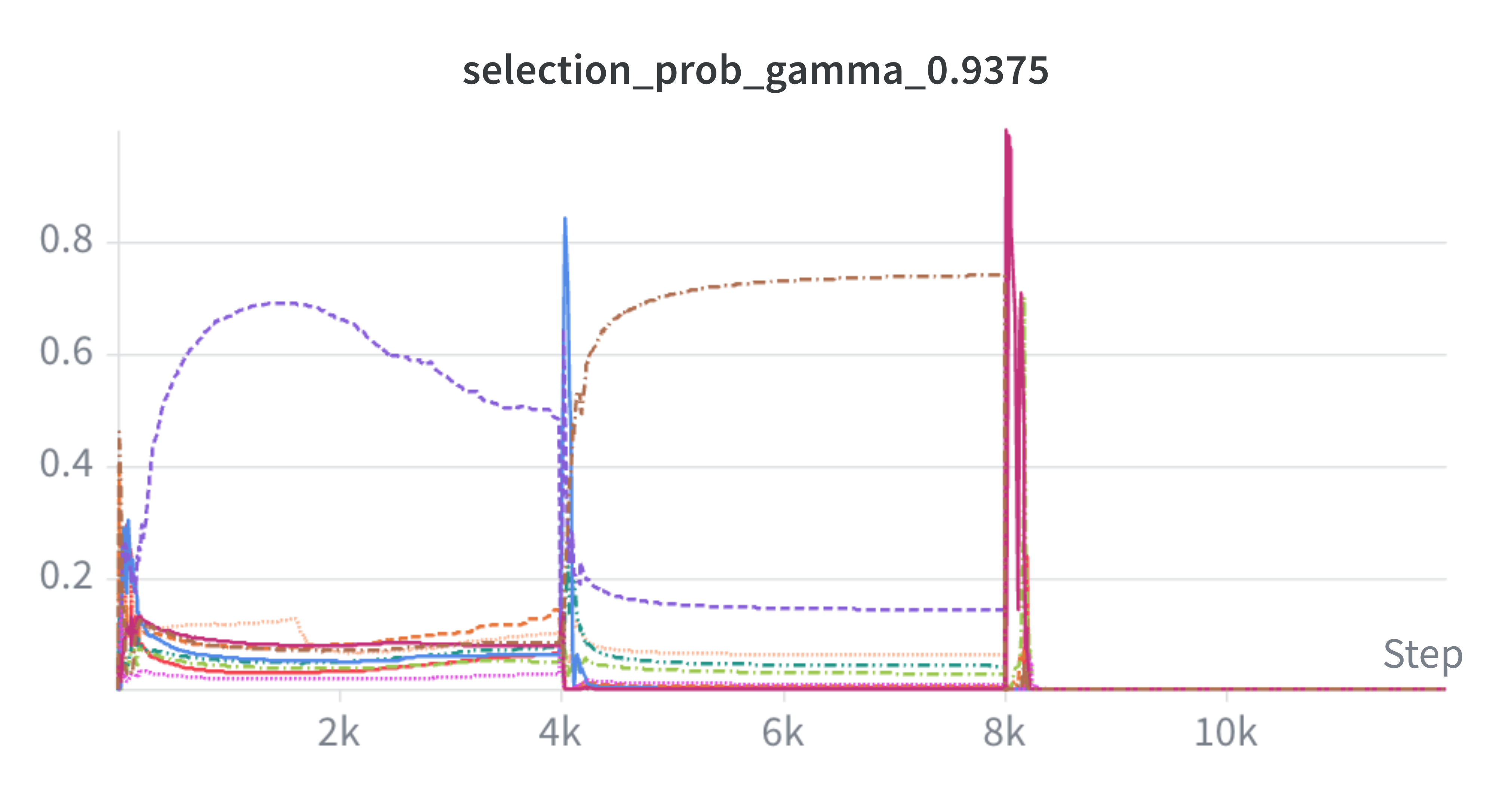}
    \end{subfigure}
    \hfill
    \begin{subfigure}[b]{0.31\textwidth}
        \centering
        \includegraphics[width=\textwidth]{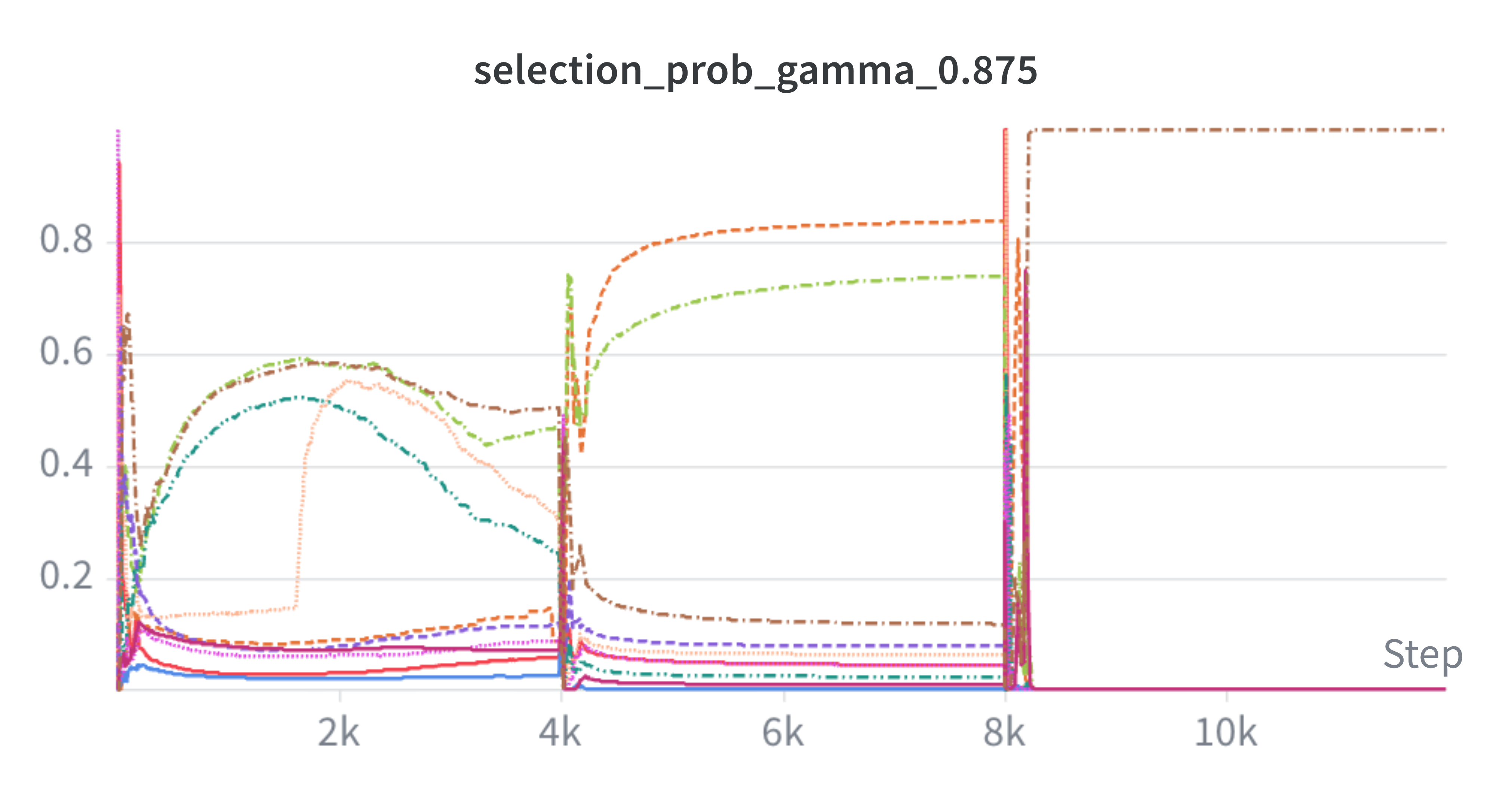}
    \end{subfigure}
    \hfill
    \begin{subfigure}[b]{0.31\textwidth}
        \centering
\includegraphics[width=\textwidth]{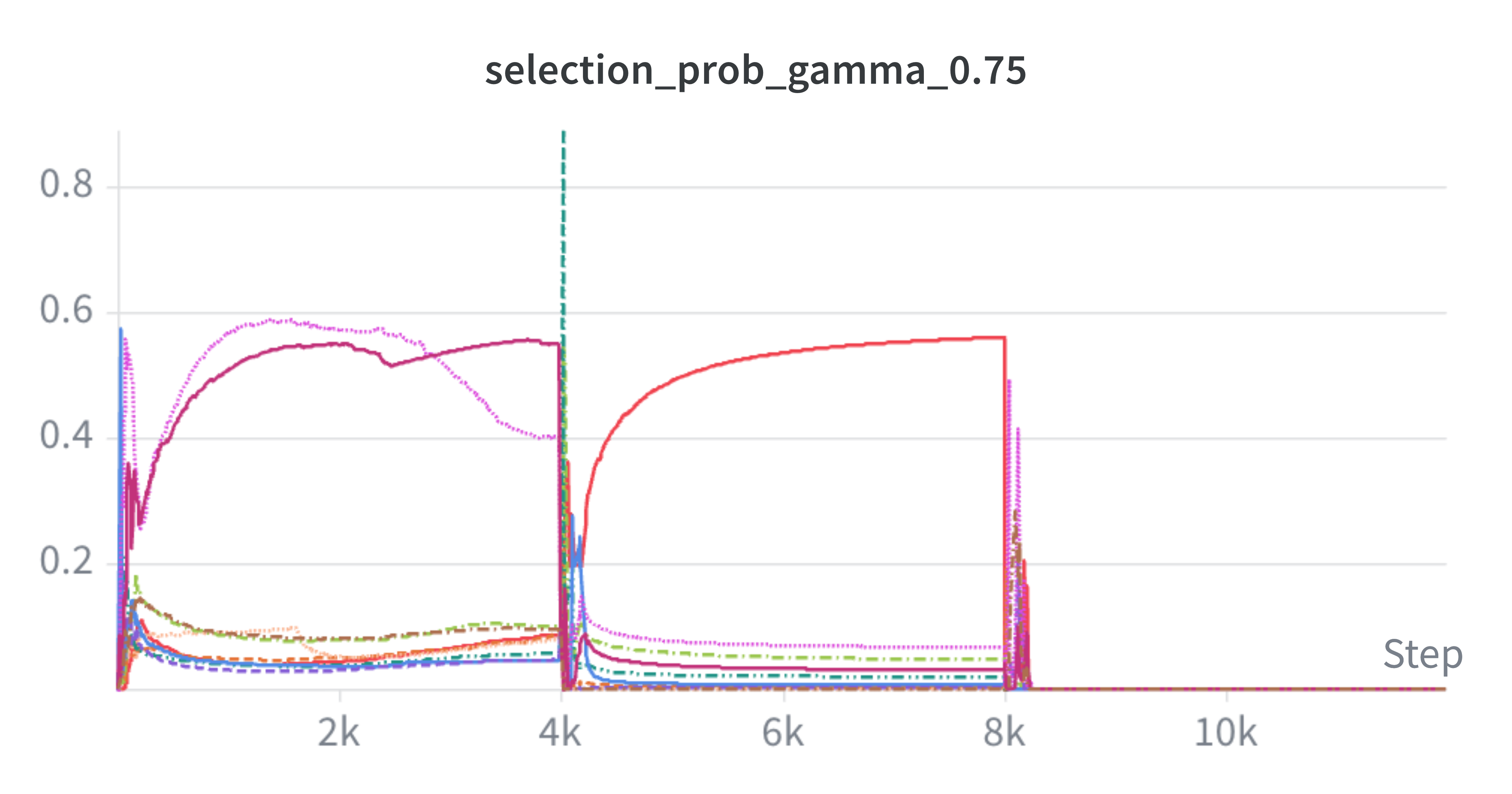}
    \end{subfigure}
    \vspace{0.2em}
    \begin{subfigure}[b]{0.31\textwidth}
        \centering
\includegraphics[width=\textwidth]{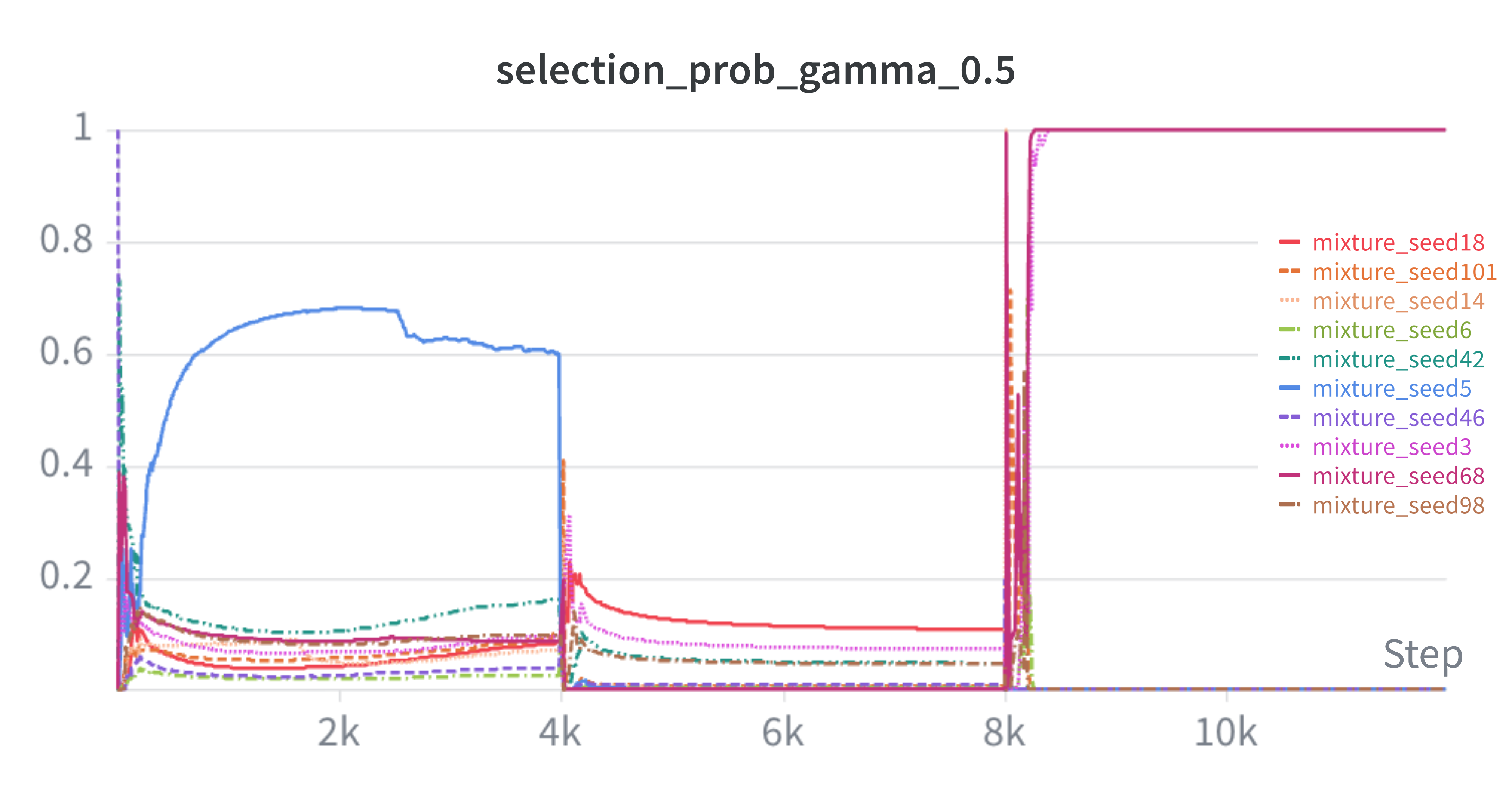}
    \end{subfigure}
    \caption{Mixer weights assigned to each discount factor during continual task switching.}
    \label{fig:weights_gamma}
\end{figure}
\vspace{-0.5em}
Although tabular RL stores values for individual state-action pairs rather than learning explicit representations, it can still exploit task structure and contextual regularities. In our environments, the first two tasks (foraging and goal reaching) represent individual components that are composed in the third task: reaching a high-value jackpot (goal) that disappears over time, followed by collecting local rewards (foraging). This compositional design allows us to study how knowledge transfers between related tasks. Similar to representation learning in neural networks, tabular agents can reuse learned state-action values and transition knowledge to adapt more efficiently when only specific task components change.

In Fig.~\ref{fig:crlreturns}, we observe that the agent efficiently adapts to each new task. The episode return remains between 36 and 40 across all runs, indicating strong performance comparable to the best individual discount-factor baselines. However, the results exhibit some sensitivity to the random seed. Evaluating the method on a larger number of seeds is an important direction for future work to better characterize and reduce this variability.
In the third task, the agent consistently learns to collect the time-constrained jackpot worth 36 points across all seeds (Fig. \ref{jackpot_}). However, collecting the remaining 4 points from the local rewards is less consistent (Fig. \ref{local_}). A likely explanation is that, after reaching the jackpot, the remaining episode length is insufficient for the agent to collect all local rewards. We leave further investigation and tuning of the episode length to future work.

Next, we examine the discount factors selected by the mixer for each task across seeds. Figure~\ref{fig:weights_gamma} shows the weights assigned to each discount factor by the mixer. For the foraging task, as previously shown in Fig.~\ref{fig:forage_return}, the best-performing discount factors are 0.9375, 0.96875, 0.875, 0.75, and 0.5. Consistent with these results, the mixer assigns most of its weight to this set of discount factors during the first 4k episodes, corresponding to the foraging task. For the goal-reaching task, as shown in Fig.~\ref{fig:goal_return}, the best-performing discount factors are all values except 0.5, 0.75, and 1.0, indicating that discount factors in the range of 0.875 to 0.998 are effective. We observe that, with the exception of two seeds (18 and 5), the mixer consistently assigns higher weights to discount factors within this optimal range. These deviations may be attributed to stochasticity in the training process. Evaluating the method over a larger number of seeds would provide a more robust assessment of this behavior. For the Four Rooms task, as shown in Fig.~\ref{fig:room_return}, the best-performing discount factors are 0.99219 and 0.98438, with 0.96875 achieving slightly lower performance. Discount factors outside this range result in substantially worse performance. Across the evaluated seeds, two seeds converge to 0.99219, one seed to 0.98438, and one seed to 0.96875. For the remaining five seeds, the mixer assigns higher weights to suboptimal discount factors, which could partially explain the lower performance in collecting local rewards. Despite this, the agent consistently learns to collect the jackpot across all seeds, including cases where a standard SARSA agent with the same discount factor fails to achieve this objective. This suggests that the learned mixture captures task-relevant structure beyond simply identifying an optimal discount factor.

\section{Conclusion and Future Work}

In summary, we believe that our mixture approach provides a novel framework for combining multiple critics with distinct capabilities, enabling adaptation to diverse and continual learning scenarios. Extending this framework beyond different discount factors, for example by incorporating critics with more diverse representations or specialized learning objectives, could further demonstrate the generality and effectiveness of the proposed methodology.

Several directions remain open for future investigation. Evaluating the mixture approach across a larger number of independent runs would provide a more comprehensive understanding of its robustness and variability. Further comparison with existing adaptive discounting methods would help contextualize its advantages and limitations. Extending the framework to jointly adapt discount factors and eligibility traces could enable more flexible control over both temporal horizons and credit assignment. Scaling beyond tabular settings through neural network function approximators or transformer-based architectures, as well as studying continual learning scenarios with naturally evolving task structures, are promising directions.

\appendix



\newpage

\section*{Acknowledgments}

We acknowledge Mila for providing access to its compute cluster and computational resources. We also acknowledge the financial support provided by UNIQUE, which contributed to making this work possible. This research was further supported by funding from the Canada First Research Excellence Fund through the Healthy Brains, Healthy Lives initiative at McGill University. Additional support for PM was provided by the Natural Sciences and Engineering Research Council of Canada (RGPIN-2025-05676 and DGECR-2025-00255), the Fonds de Recherche du Québec (CB-365865), and a Sloan Research Fellowship. This research was enabled in part by computational resources provided by Calcul Québec and the Digital Research Alliance of Canada.


\bibliography{main}
\bibliographystyle{rlj}

\section*{Appendix}

\subsection*{Gradient Derivation of Multi-Timescale Mixture of Q-Values}
\label{app:derivation}
The mixed action-value function is defined as
\begin{equation}
Q_{\mathrm{mix}}(s,a)
=
\sum_{i=1}^{K}
w_i(s)Q_i(s,a).
\end{equation}
$Q_{\mathrm{mix}}$ follows an expected SARSA with $\gamma=1$, the TD error is
\begin{equation}
\delta
=
r_{t+1}
+
\sum_{a'}
\pi(a'|s_{t+1})
Q_{\mathrm{mix}}(s_{t+1},a')
-
Q_{\mathrm{mix}}(s_t,a_t).
\label{undiscounted_TD}
\end{equation}
The gating network parameters are optimized by minimizing the squared TD error,
\begin{equation}
\mathcal{L}
=
\frac{1}{2}\delta^2 .
\end{equation}
The TD error can be written as
\begin{equation}
\delta
=
y-Q_{\mathrm{mix}}(s,a),
\end{equation}
where
\begin{equation}
y=
r+
\sum_{a'}
\pi(a'|s')
Q_{\mathrm{mix}}(s',a')
\end{equation}
is treated as a constant during the semi-gradient update. Therefore,
\begin{equation}
\frac{\partial \delta}
{\partial Q_{\mathrm{mix}}}
=
-1 .
\end{equation}
Using the chain rule,
\begin{equation}
\frac{\partial \mathcal{L}}
{\partial Q_{\mathrm{mix}}}
=
\frac{\partial \mathcal{L}}
{\partial \delta}
\frac{\partial \delta}
{\partial Q_{\mathrm{mix}}}.
\end{equation}
Since
\begin{equation}
\frac{\partial \mathcal{L}}
{\partial \delta}
=
\delta,
\end{equation}
we obtain
\begin{equation}
\frac{\partial \mathcal{L}}
{\partial Q_{\mathrm{mix}}}
=
-\delta .
\end{equation}
Next, we compute the derivative of the mixture value with respect to the
logit of expert $i$. The softmax derivative is
\begin{equation}
\frac{\partial w_k}{\partial z_i}
=
w_k(\mathbb{I}_{ki}-w_i),
\end{equation}
where $\mathbb{I}_{ki}$ is the Kronecker delta.
The derivative of the mixed value is therefore
\begin{equation}
\frac{\partial Q_{\mathrm{mix}}}
{\partial z_i}
=
\sum_{k=1}^{K}
Q_k
\frac{\partial w_k}
{\partial z_i}.
\end{equation}
Substituting the softmax derivative,
\begin{equation}
\frac{\partial Q_{\mathrm{mix}}}
{\partial z_i}
=
\sum_{k=1}^{K}
Q_k w_k(\mathbb{I}_{ki}-w_i).
\end{equation}
Separating the two terms gives
\begin{equation}
\frac{\partial Q_{\mathrm{mix}}}
{\partial z_i}
=
w_iQ_i
-
w_i
\sum_{k=1}^{K}
w_kQ_k .
\end{equation}
Since
\begin{equation}
Q_{\mathrm{mix}}
=
\sum_{k=1}^{K}
w_kQ_k,
\end{equation}
we obtain
\begin{equation}
\frac{\partial Q_{\mathrm{mix}}}
{\partial z_i}
=
w_i(Q_i-Q_{\mathrm{mix}}).
\end{equation}
The term
\begin{equation}
A_i
=
Q_i-Q_{\mathrm{mix}}
\end{equation}
represents the advantage of expert $i$ relative to the current mixture.
Using
\begin{equation}
\frac{\partial z_i}{\partial W_i}=x,
\end{equation}
the gradient of the loss with respect to the gating parameters is
\begin{equation}
\frac{\partial \mathcal{L}}
{\partial W}
=
\frac{\partial \mathcal{L}}
{\partial Q_{\mathrm{mix}}}
\frac{\partial Q_{\mathrm{mix}}}
{\partial z}
\frac{\partial z}
{\partial W}.
\end{equation}
Substituting the previous results,
\begin{equation}
\frac{\partial \mathcal{L}}
{\partial W}
=
-\delta
\left[
w\odot(Q-Q_{\mathrm{mix}})
\right]
x^\top ,
\end{equation}
where $\odot$ denotes element-wise multiplication.
Finally, applying gradient descent,
\begin{equation}
W
\leftarrow
W-\alpha
\frac{\partial \mathcal{L}}
{\partial W},
\end{equation}
which gives
\begin{equation}
W
\leftarrow
W+
\alpha\delta
\left[
w\odot(Q-Q_{\mathrm{mix}})
\right]
x^\top .
\end{equation}
The bias parameters are updated similarly,
\begin{equation}
b
\leftarrow
b+
\alpha\delta
\left[
w\odot(Q-Q_{\mathrm{mix}})
\right].
\end{equation}
\subsection*{Exploratory Theoretical Analysis}

\begin{proposition}[Convex Representation of Horizon-Dependent Critics]
The convex hull of a set of functions
$\{Q_1,\dots,Q_K\}$ is defined as the set of all convex combinations:
\begin{equation}
\mathrm{conv}\{Q_1,\dots,Q_K\}
=
\left\{
\sum_{i=1}^{K}w_iQ_i
\;\middle|\;
w_i\geq0,\;
\sum_{i=1}^{K}w_i=1
\right\}.
\end{equation}
Since the gating network produces normalized mixture weights satisfying
\begin{equation}
w_i(s)\geq0,
\qquad
\sum_{i=1}^{K}w_i(s)=1,
\end{equation}
the constraints of a convex combination. 
Therefore, for every state-action pair $(s,a)$, $Q_w$ lies in the convex hull of the horizon-dependent critics ($
Q_w(s,a)
\in
\mathrm{conv}
\{Q_1(s,a),\dots,Q_K(s,a)\}
$), that best approximates the full episodic return objective, allowing the agent to smoothly interpolate between different temporal
abstractions rather than selecting a single fixed discount factor.
\end{proposition}

\begin{proposition}[Undiscounted Bellman Residual of the Mixture Critic]

We optimize the undiscounted episodic return objective
\begin{equation}
J(\pi_w)
=
\mathbb{E}
\left[
\sum_{t=0}^{T}
r_t
\right],
\end{equation}

which maximizes the total reward collected over an episode. For a fixed policy, the undiscounted Bellman operator is defined as:
\begin{equation}
T_{1}Q(s,a)
=
r(s,a)
+
\sum_{a'}
\pi(a'|s')
Q(s',a').
\end{equation}

The corresponding TD error is given in
Equation~\ref{undiscounted_TD}, which can be written as the Bellman residual:
\begin{equation}
\delta_t
=
T_{1}Q_w(s_t,a_t)
-
Q_w(s_t,a_t).
\end{equation}

Thus, the semi-gradient TD update can be interpreted as minimizing the
undiscounted Bellman error:
\begin{equation}
\mathcal{E}(w)
=
\frac{1}{2}
\mathbb{E}
\left[
\delta_t^2
\right]
=
\frac{1}{2}
\mathbb{E}
\left[
\left(
T_{1}Q_w
-
Q_w
\right)^2
\right].
\end{equation}

Because $Q_w$ is restricted to the convex hull of the horizon-dependent
critics, minimizing $\mathcal{E}(w)$ finds the convex combination of temporal
abstractions that best satisfies the undiscounted Bellman equation.
\end{proposition}

\begin{theorem}[Sufficient Conditions for Bellman-Residual-Based Horizon Selection]

We assume

\begin{enumerate}
\item Each expert converges to its corresponding discounted action-value
function:
\begin{equation}
Q_i
\rightarrow
Q_{\gamma_i}^{\pi}.
\end{equation}
\item There exists an optimal temporal horizon $\gamma^*$ satisfying
\begin{equation}
\gamma^*
=
\arg\max_{\gamma_i}
J(\pi_{\gamma_i}),
\end{equation}
where
\begin{equation}
J(\pi)
=
\mathbb{E}
\left[
\sum_{t=0}^{T}
r_t
\right]
\end{equation}
denotes the undiscounted episodic return.
\item The critic associated with the optimal horizon provides the most
accurate approximation to the undiscounted Bellman equation:
\begin{equation}
\mathbb{E}
\left[
\delta_{\gamma^*}^{2}
\right]
<
\mathbb{E}
\left[
\delta_{\gamma_i}^{2}
\right],
\qquad
\forall i\neq * ,
\end{equation}
where
\begin{equation}
\delta_{\gamma_i}
=
T_{1}Q_{\gamma_i}
-
Q_{\gamma_i}.
\end{equation}
\item Bellman residuals from different experts do not cancel in the convex combination.

\end{enumerate}

Under these assumptions, minimizing the undiscounted Bellman residual over the
mixture weights assigns increasing mass to the critic with the smallest
Bellman residual:
\begin{equation}
w_{\gamma^*}\rightarrow1,
\qquad
w_i\rightarrow0,
\quad
i\neq * .
\end{equation}

Therefore, minimizing the undiscounted Bellman error recovers the temporal
horizon that maximizes the undiscounted episodic return:
\begin{equation}
\boxed{
\gamma^*
=
\arg\max_{\gamma_i}
J(\pi_{\gamma_i})
}.
\end{equation}

Thus, the convex mixture formulation enables adaptive temporal abstraction:
rather than committing to a single discount factor, the agent learns a
state-dependent combination of horizon-dependent critics that best
approximates the undiscounted return objective.

\end{theorem}

\subsection*{Figures}
\begin{figure}[htbp!]
    \centering
    
    \begin{subfigure}[b]{0.49\textwidth}
        \centering
     \includegraphics[width=\textwidth]{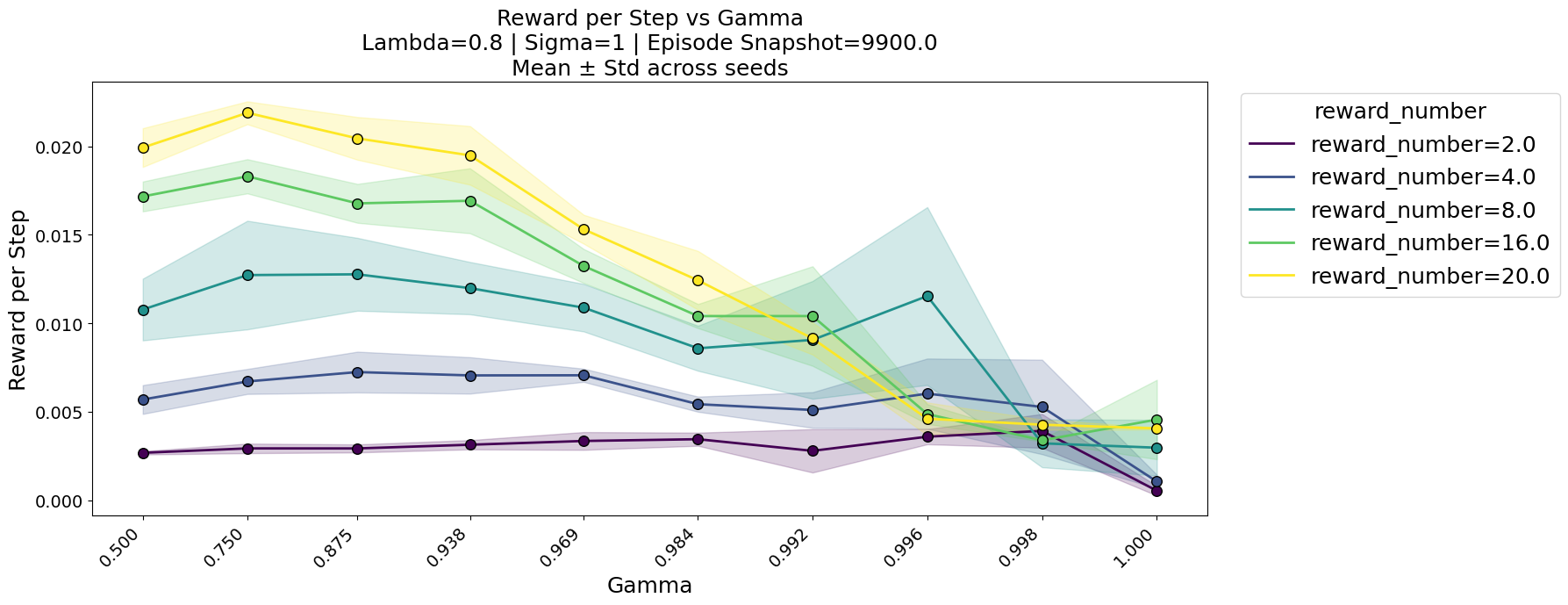}
        \caption{Reward per step vs. $\gamma$.}
        \label{fig:reward_per_step_sig1}
    \end{subfigure}
    \hfill
    \begin{subfigure}[b]{0.49\textwidth}
        \centering
        \includegraphics[width=\textwidth]{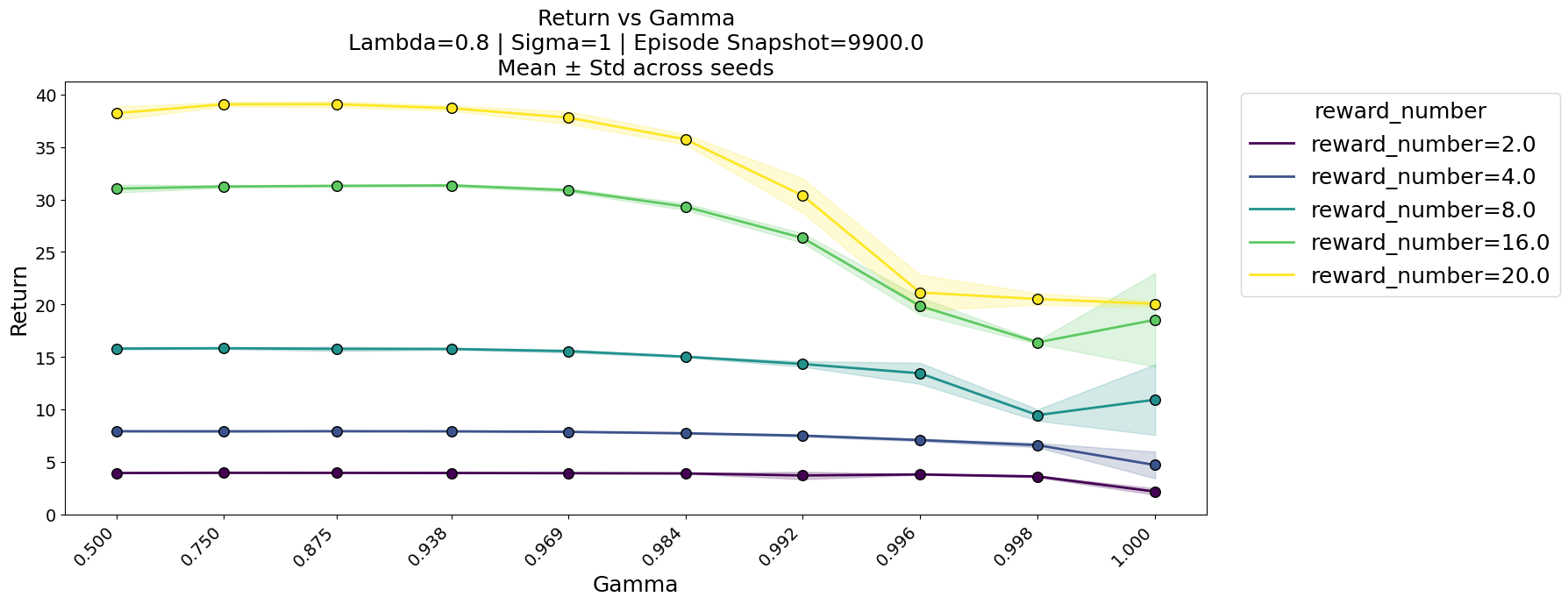}
        \caption{Return vs. $\gamma$.}
        \label{fig:return_vs_gamma_sig1}
    \end{subfigure}
    \caption{Return and reward per step, as a function of the discount factor ($\gamma$) for different numbers of rewards per cluster (${2, 4, 8, 16, 20}$), averaged over 5 seeds. The trace parameter ($\lambda$) is fixed at 0.8. Gaussian standard deviation ($\sigma$) is fixed at 1. }
    \label{fig:gamma_freq_rew_sig1}
\end{figure}

\begin{figure}[ht!]
    \begin{subfigure}[b]{0.31\textwidth}
        \centering
    \includegraphics[width=0.55\textwidth]{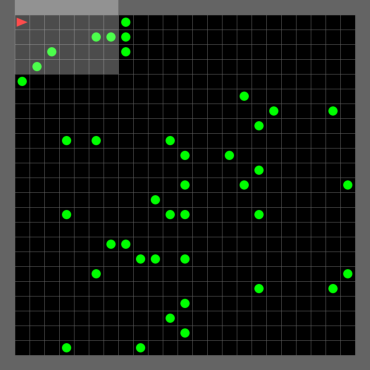}
        \caption{Task 1: Foraging}
        \label{fig:forage}
    \end{subfigure}
        \hfill
    \begin{subfigure}[b]{0.31\textwidth}
        \centering \includegraphics[width=0.55\textwidth]{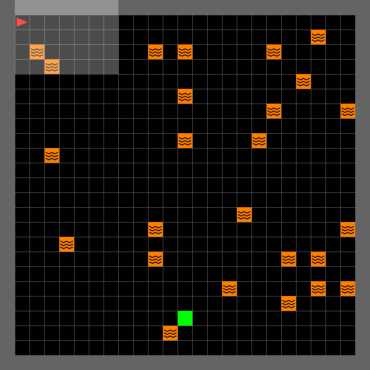}
        \caption{Task 2: Goal Reaching}
        \label{fig:goal}
    \end{subfigure}
    \hfill
    \begin{subfigure}[b]{0.31\textwidth}
        \centering
\includegraphics[width=0.55\textwidth]{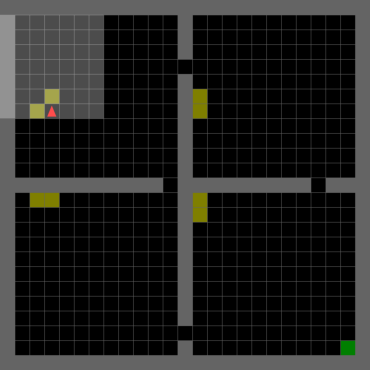}
        \caption{Task 3: Four Rooms}
        \label{fig:room}
    \end{subfigure}
    \caption{The three tasks in the continual setup, presented to the agent in sequential order.}
    \label{fig:continual}
\end{figure}

\begin{figure}[htbp!]
\centering
\includegraphics[width=0.7\textwidth]{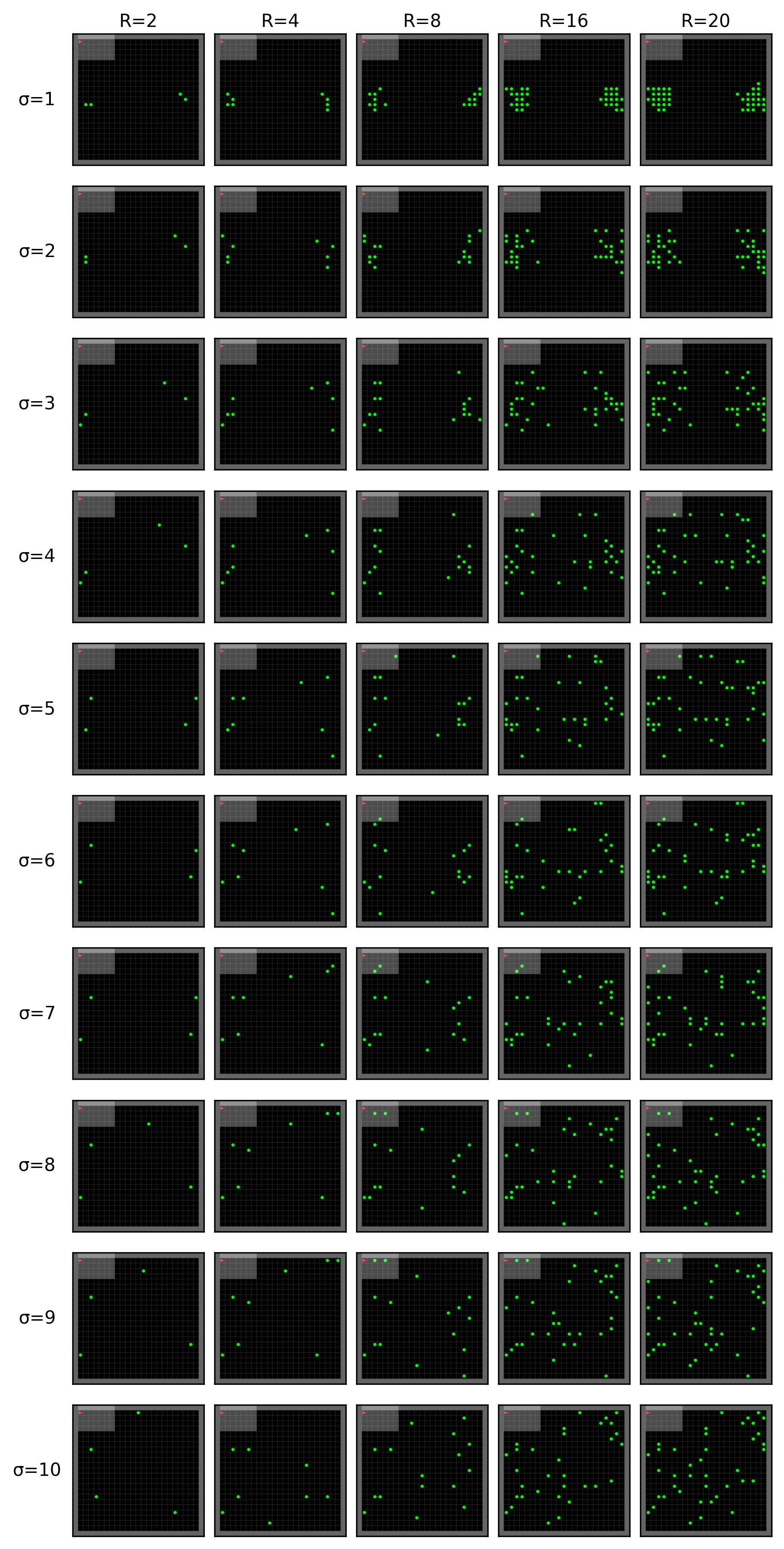}
        
\caption{Grid layout with increasing Gaussian sampling variance (rows) and number of rewards (columns).}
\label{fig:sigma-num}
\end{figure}



%

\end{document}